\documentclass{article}

\usepackage{hyperref} %[draft]
\usepackage[accepted]{icml2020}

\usepackage{natbib}
\usepackage{graphicx}
\usepackage[utf8]{inputenc}
\usepackage{amssymb,amsthm,array,amsfonts}
\usepackage{color}
\usepackage{booktabs}
\usepackage{underoverlap}
\usepackage{subfigure}
\usepackage{multirow}
\usepackage{hyperref}
\usepackage{mathtools}% Loads amsmath
\usepackage[rightcaption]{sidecap}
\sidecaptionvpos{figure}{c}
\usepackage{appendix}

\def\x{{\mathbf x}}

\def\y{{\mathbf y}}

\def\s{{\mathbf s}}
\def\X{{\mathbf X}}

\def\U{{\mathbf U}}

\def\C{{\mathbf C}}

\def\A{{\mathbf A}}

\def\S{{\mathbf S}}
\def\D{{\mathbf D}}
\def\I{{\mathbf I}}
\def\Q{{\mathbf Q}}
\def\R{{\mathbb R}}

\def\GNN{{\textrm{GNN}}}
\def\MLP{{\textrm{MLP}}}

  

\newcommand{\mincut}{\texttt{\small{minCUT}}}
\newcommand{\mincutpool}{MinCutPool}
\renewcommand{\t}[1]{\tiny{#1}}

\newcommand{\argmax}{\operatornamewithlimits{arg\ max}}

\icmltitlerunning{Spectral Clustering with Graph Neural Networks for Graph Pooling}

\begin{document}

\twocolumn[
\icmltitle{Spectral Clustering with Graph Neural Networks for Graph Pooling}



\icmlsetsymbol{equal}{*}

\begin{icmlauthorlist}
\icmlauthor{Filippo Maria Bianchi}{equal,nor}
\icmlauthor{Daniele Grattarola}{equal,usi}
\icmlauthor{Cesare Alippi}{usi,poli}
\end{icmlauthorlist}

\icmlaffiliation{nor}{NORCE, the Norwegian Research Centre, Norway}
\icmlaffiliation{usi}{Faculty of Informatics, Universit\`a della Svizzera italiana, Lugano, Switzerland}
\icmlaffiliation{poli}{DEIB, Politecnico di Milano, Milano, Italy}

\icmlcorrespondingauthor{Filippo Maria Bianchi}{filippombianchi@gmail.com}

\icmlkeywords{Graph neural networks, graph pooling, graph coarsening}

\vskip 0.3in
]

\printAffiliationsAndNotice{\icmlEqualContribution} % otherwise use the standard text.

\begin{abstract}
Spectral clustering (SC) is a popular clustering technique to find strongly connected communities on a graph.
SC can be used in Graph Neural Networks (GNNs) to implement pooling operations that aggregate nodes belonging to the same cluster.
However, the eigendecomposition of the Laplacian is expensive and, since clustering results are graph-specific, pooling methods based on SC must perform a new optimization for each new sample.
In this paper, we propose a graph clustering approach that addresses these limitations of SC.
We formulate a continuous relaxation of the normalized \mincut{} problem and train a GNN to compute cluster assignments that minimize this objective. 
Our GNN-based implementation is differentiable, does not require to compute the spectral decomposition, and learns a clustering function that can be quickly evaluated on out-of-sample graphs.
From the proposed clustering method, we design a graph pooling operator that overcomes some important limitations of state-of-the-art graph pooling techniques and achieves the best performance in several supervised and unsupervised tasks.
\end{abstract}

%%%%%%%%%%%%%%%%%%%%%%%%%%%%%%%%%%%%%%%%%%%%%%%%%%
% INTRODUCTION
%%%%%%%%%%%%%%%%%%%%%%%%%%%%%%%%%%%%%%%%%%%%%%%%%%
\section{Introduction}

State-of-the-art convolutional neural networks (CNNs) alternate convolutions, which extrapolate local features from the input signal, with pooling, which downsamples the feature maps by computing local summaries of nearby points. 
Pooling helps CNNs to discard information that is superfluous for the task, provides translation invariance, and keeps model complexity under control by reducing the size of the intermediate representations.

\begin{figure}
    \centering
    \includegraphics[width=.9\columnwidth]{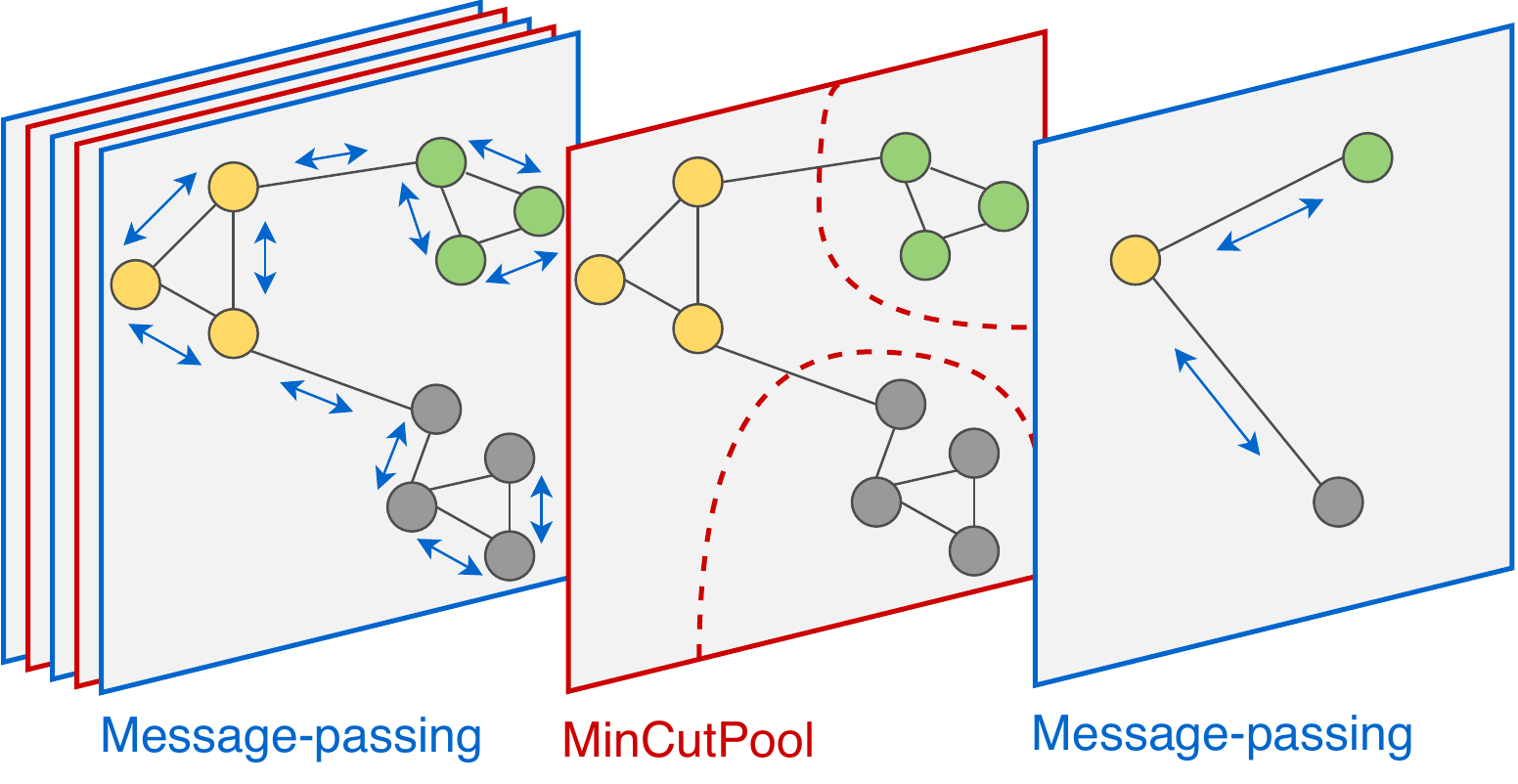}
    \caption{A deep GNN architecture where message-passing is followed by the \mincutpool{} layer.}
    \label{fig:gnn_pool}
\end{figure}

Graph Neural Networks (GNNs) extend the convolution operation from regular domains to arbitrary topologies and unordered structures~\citep{battaglia2018relational}.
As in CNNs, graph pooling is an important operation that allows a GNN to learn increasingly more abstract and coarser representations of the input graphs, by summarizing local components and discarding redundant information.
The development of pooling strategies for GNNs, however, has lagged behind the design of newer and more effective message-passing (MP) operations~\citep{gilmer2017neural}, such as graph convolutions.
The reason, is mainly due to the difficulty of defining an aggregated version of a graph that effectively supports the pooled node features. 

Several approaches have been proposed in recent GNN literature, ranging from model-free methods that pre-compute the pooled graphs by leveraging graph-theoretical properties~\citep{bruna2013spectral,defferrard2016convolutional}, to model-based methods that perform pooling trough a learnable function of the node features~\citep{ying2018hierarchical,cangea2018towards,graphunet}. 
However, existing model-free methods only consider the graph topology but ignore the node features, while model-based methods are mostly based on heuristics. 
As a consequence, the former cannot learn how to coarsen graphs adaptively for a specific downstream task, while the latter are unstable and prone to find degenerate solutions even on simple tasks.
A pooling operation that is theoretically grounded and can adapt to the data and the task is still missing.

Spectral clustering (SC) is a well-known clustering technique that leverages the Laplacian spectrum to find strongly connected communities on a graph. 
SC can be used to perform pooling in GNNs by aggregating nodes belonging to the same cluster~\citep{bruna2013spectral,defferrard2016convolutional}, although the approaches based on this technique suffer from the aforementioned issues of model-free pooling methods. 
In particular, SC does not explicitly account for the node attributes and the eigendecomposition of the Laplacian is a non-differentiable and expensive operation.
Additionally, pooling methods based on SC must compute the spectral decomposition even at test time, since the decomposition is unique for each new graph.

We propose a graph clustering approach that addresses the limitations that hinder the applicability of SC in GNNs.
Specifically, we formulate a continuous relaxation of the normalized \mincut{} problem and train a GNN to compute cluster assignments by optimizing this objective.
Our approach learns the solution found by SC while also accounting explicitly for the node features to identify clusters.
At the same time, our GNN-based implementation is differentiable and does not require to compute the expensive spectral decomposition of the Laplacian, since it exploits spatially localized graph convolutions that are fast to compute.
This also allows to cluster the nodes of out-of-sample graphs simply by evaluating the learned function.

From the proposed clustering method, we derive a model-based pooling operator called \textit{\mincutpool{}}, which overcomes the disadvantages of both model-free and model-based pooling methods.
The parameters in a \mincutpool{} layer are learned by minimizing the \mincut{} objective, which can be jointly optimized with a task-specific loss. 
In the latter case, the \mincut{} loss acts as a regularization term, which prevents degenerate solutions, and the GNN can find the optimal trade-off between task-specific and clustering objectives.
Because they are fully differentiable, \mincutpool{} layers can be stacked at different levels of a GNN to obtain a hierarchical representation and the overall architecture can be trained end-to-end (Figure \ref{fig:gnn_pool}). 
We perform a comparative study on a variety of unsupervised and supervised tasks and show that \mincutpool{} leads to significant improvements over state-of-the-art pooling methods.

%%%%%%%%%%%%%%%%%%%%%%%%%%%%%%%%%%%%%%%%%%%%%%%%%%
% BACKGROUND
%%%%%%%%%%%%%%%%%%%%%%%%%%%%%%%%%%%%%%%%%%%%%%%%%%
\section{Background}
Let a graph be represented by a tuple $G = \{ \mathcal{V}, \mathcal{E} \}$, $|\mathcal{V}|=N$, with node set $\mathcal{V}$ and edge set $\mathcal{E}$. Each node is associated with a vector attribute in $\mathbb{R}^F$.
A graph is characterized by its adjacency matrix $\A \in \mathbb{R}^{N \times N}$ and the node features $\X \in \R^{N \times F}$.

\subsection{Graph Neural Networks}
Several approaches have been proposed to process graphs with neural networks, including recurrent architectures~\citep{scarselli2009graph, li2015gated, gallicchio2019fast} or convolutional operations inspired by filters used in graph signal processing~\citep{defferrard2016convolutional,kipf2016semi,bianchi2019graph}.
We base our GNN architecture on a simple MP operation that combines the features of each node with its first-order neighbours. We adopt a MP implementation that does not require to modify the graph by adding self-loops (like in, e.g., \cite{kipf2016semi}) but accounts for the initial node features through a skip connection.

Let $\tilde{\A} = \D^{-\frac{1}{2}}\A\D^{-\frac{1}{2}} \in \R^{N \times N}$ be the symmetrically normalized adjacency matrix, where $\D = \text{diag}(\A \boldsymbol{1}_N)$ is the degree matrix.
The output of the MP layer is
\begin{equation}
    \label{eq:mp}
    \bar \X = MP(\X, \tilde \A) = \text{ReLU}(\tilde \A \X \boldsymbol{\Theta}_m + \X \boldsymbol{\Theta}_s),
\end{equation}
where $\boldsymbol{\Theta}_m$ and $\boldsymbol{\Theta}_s$ are the trainable weights of the mixing and skip components of the layer, respectively.

\subsection{\mincut{} and Spectral Clustering}
\label{sec:mincut}
Given a graph $G = \{ \mathcal{V}, \mathcal{E} \}$, the \textit{$K$-way normalized} \mincut{} problem (simply referred to as \mincut{}) is the task of partitioning $\mathcal{V}$ in $K$ disjoint subsets by removing the minimum volume of edges. 
The problem is equivalent to maximizing
\begin{equation}
    \label{eq:ncut}
    \frac{1}{K} \sum_{k=1}^K \frac{\text{links}(\mathcal{V}_k)}{\text{degree}(\mathcal{V}_k)} = \frac{1}{K} \sum_{k=1}^K \frac{\sum_{i,j \in \mathcal{V}_k} \mathcal{E}_{i,j} }{\sum_{i \in \mathcal{V}_k, j \in \mathcal{V} \backslash \mathcal{V}_k} \mathcal{E}_{i,j}},
\end{equation}
where the numerator counts the edge volume within each cluster, and the denominator counts the edges between the nodes in a cluster and the rest of the graph~\citep{shi2000normalized}.
Let $\C \in \{0,1\}^{N \times K}$ be a \textit{cluster assignment matrix}, so that $\C_{i,j} = 1$ if node $i$ belongs to cluster $j$, and 0 otherwise.
The \mincut{} problem can be expressed as~\citep{dhillon2004kernel}:
\begin{equation}
\label{eq:ncut_optim}
    \begin{aligned}
        & \text{maximize} \;\; \frac{1}{K} \sum_{k=1}^K \frac{\C_k^T \A \C_k}{\C_k^T \D \C_k}, \\
        & \text{s.t.} \;\; \C \in \{0,1\}^{N \times K}, \;\;
        \C \boldsymbol{1}_K = \boldsymbol{1}_N
    \end{aligned}
\end{equation}
where $\C_k$ is the $k$-th column of $\C$.
Since problem \eqref{eq:ncut_optim} is NP-hard, it is recast in a relaxed continuous formulation that can be solved in polynomial time and guarantees a near-optimal solution~\citep{1238361}: 
\begin{equation}
\label{eq:ncut_relax}
    \begin{aligned}
        & \argmax_{\Q \in \mathbb{R}^{N \times K}} \;\; \frac{1}{K} \sum_{k=1}^K \Q_k^T \A \Q_k \\
        & \text{s.t.} \; \Q = \D^{\frac{1}{2}}\C(\C^T\D\C)^{-\frac{1}{2}},\; \Q^T \Q = \I_K.
    \end{aligned}
\end{equation}
While problem \eqref{eq:ncut_relax} is still non-convex, there exists an optimal solution $\Q^{*} = \U_K \mathbf{O}$, where $\U_K \in \mathbb{R}^{N \times K}$ contains the eigenvectors of $\tilde \A$ corresponding to the $K$ largest eigenvalues, and $\mathbf{O} \in \mathbb{R}^{K \times K}$ is an orthogonal transformation~\citep{ikebe1987monotonicity}.

Spectral clustering (SC) obtains the cluster assignments by applying $k$-means to the rows of $\Q^{*}$, which are node embeddings in the Laplacian eigenspace~\citep{von2007tutorial}.
One of the main limitations of SC lies in the computation of the spectrum of $\tilde \A$ to obtain $\Q^{*}$, which has a memory complexity of $\mathcal{O}(N^2)$ and a computational complexity of $\mathcal{O}(N^3)$. 
This prevents its applicability to large datasets.

To deal with the scalability issues of SC, the constrained optimization in \eqref{eq:ncut_relax} can be solved by gradient descent algorithms that refine the solution by iterating operations whose individual complexity is $\mathcal{O}(N^2)$, or even $\mathcal{O}(N)$~\citep{han2017mini}.
These algorithms search the solution on the manifold induced by the orthogonality constraint on the columns of $\Q$, by performing gradient updates along the geodesics~\citep{wen2013feasible, collins2014spectral}.
Alternative approaches rely on QR factorization to constrain the space of feasible solutions~\citep{damle2016robust}, and alleviate the cost $\mathcal{O}(N^3)$ of the factorization by ensuring that orthogonality holds only on one minibatch at a time~\citep{shaham2018spectralnet}.
\citet{dhillon2007weighted} discuss the equivalence between graph clustering objectives and the kernel k-means algorithm, and their Graclus algorithm is a popular model-free method for hierarchical pooling in GNNs~\citep{defferrard2016convolutional}.

To learn a model that finds an approximate SC solution also for out-of-sample graphs, several works propose to use neural networks. 
In \citep{tian2014learning}, an autoencoder is trained to map the $i^\text{th}$ row of the Laplacian to the $i^\text{th}$ components of the first $K$ eigenvectors.
\citet{yi2017syncspeccnn} define an orthogonality constraint to learn spectral embeddings as a volumetric reparametrization of a precomputed Laplacian eigenbasis. Finally, \citet{shaham2018spectralnet} propose a loss function to cluster generic data and process out-of-sample data at inference time. 
While these approaches learn to embed data in the Laplacian eigenspace of the given graph, they rely on non-differentiable operations to compute the cluster assignments and, therefore, are not suitable to perform pooling in a GNN trained end-to-end.

%%%%%%%%%%%%%%%%%%%%%%%%%%%%%%%%%%%%%%%%%%%%%%%%%%
% METHOD
%%%%%%%%%%%%%%%%%%%%%%%%%%%%%%%%%%%%%%%%%%%%%%%%%%
\section{Spectral Clustering with GNNs}
\label{sec:method}

We propose a GNN-based approach that addresses the aforementioned limitations of SC algorithms and that clusters the nodes according to the graph topology (nodes in the same cluster should be strongly connected) and to the node features (nodes in the same cluster should have similar features).
Our method assumes that node features represent a good initialization for computing the cluster assignments. 
This is a realistic assumption due to the homophily property of many real-world networks~\cite{mcpherson2001birds}.
Additionally, even in disassortative networks (i.e., networks where dissimilar nodes are likely to be connected~\citep{newman2003mixing}), the features of nodes in strongly connected communities tend to become similar due to the smoothing effect of MP operation.

Let $\bar \X$ be the matrix of node representations yielded by one or more MP layers. 
We compute a cluster assignment of the nodes using a multi-layer perceptron (MLP) with softmax on the output layer, which maps each node feature $\x_i$ into the $i^\text{th}$ row of a soft cluster assignment matrix $\S$:
\begin{equation}
    \label{eq:assigments}
    \begin{aligned}
        \bar \X & = \GNN(\X, \tilde \A; \boldsymbol{\Theta}_{\GNN}) \\
        \S & = \MLP(\bar \X; \boldsymbol{\Theta}_{\MLP}),
    \end{aligned}
\end{equation}
where $\boldsymbol{\Theta}_{\GNN}$ and $\boldsymbol{\Theta}_{\MLP}$ are trainable parameters.
The softmax activation of the MLP guarantees that $\s_{ij} \in [0,1]$ and enforces the constraints $\S\boldsymbol{1}_K = \boldsymbol{1}_N$ inherited from the optimization problem in \eqref{eq:ncut_optim}.
We note that it is possible to add a temperature parameter to the Softmax in the MLP to control how much $\mathbf{s}_i$ should be close to a one-hot vector, \textit{i.e.}, the level of fuzziness in the cluster assignments.

The parameters $\boldsymbol{\Theta}_{\GNN}$ and $\boldsymbol{\Theta}_{\MLP}$ are jointly optimized by minimizing an unsupervised loss function $\mathcal{L}_u$ composed of two terms, which approximates the relaxed formulation of the \mincut{} problem:
\begin{equation}
    \label{eq:norm_cut}
    \mathcal{L}_u = \mathcal{L}_c + \mathcal{L}_o = 
    \underbrace{- \frac{Tr ( \S^T \tilde \A \S )}{Tr ( \S^T \tilde \D \S)}}_{\mathcal{L}_c} + 
    \underbrace{\bigg{\lVert} \frac{\S^T\S}{\|\S^T\S\|_F} - \frac{\I_K}{\sqrt{K}}\bigg{\rVert}_F}_{\mathcal{L}_o}, 
\end{equation}
where $\| \cdot \|_F$ indicates the Frobenius norm and $\tilde \D$ is the degree matrix of $\tilde \A$.

The \textit{cut loss} term, $\mathcal{L}_c$, evaluates the \mincut{} given by the soft cluster assignment $\S$ and is bounded by $-1 \leq \mathcal{L}_c \leq 0$.
Minimizing $\mathcal{L}_c$ encourages strongly connected nodes to be clustered together, since the inner product $\langle \s_i, \s_j \rangle$ increases when $\tilde{a}_{i,j}$ is large.
$\mathcal{L}_c$ has a single maximum, reached when the numerator $Tr ( \S^T \tilde \A \S ) = \frac{1}{K} \sum_{k=1}^K \S_k^T \tilde \A \S_k=0$. This occurs if, for each pair of connected nodes (i.e., $\tilde a_{i,j} > 0$), the cluster assignments are orthogonal (i.e., $\langle \s_i, \s_j \rangle = 0$).
$\mathcal{L}_c$ reaches its minimum, $-1$, when $Tr( \S^T \tilde \A \S ) = Tr( \S^T \tilde \D \S )$.
This occurs when, in a graph with $K$ disconnected components, the cluster assignments are equal for all the nodes in the same component and orthogonal to the cluster assignments of nodes in different components.
However, $\mathcal{L}_c$ is a non-convex function and its minimization can lead to local minima or degenerate solutions.
For example, given a connected graph, a trivial - yet optimal - solution is the one that assigns all nodes to the same cluster.
As a consequence of the continuous relaxation, another degenerate minimum occurs when the cluster assignments are all uniform, that is, all nodes are equally assigned to all clusters.
This problem is exacerbated by the MP operations, whose smoothing effect makes the node features more uniform.

To penalize the degenerate minima of $\mathcal{L}_c$, the orthogonality loss term $\mathcal{L}_o$ encourages the cluster assignments to be orthogonal and the clusters to be of similar size.
Since the two matrices in $\mathcal{L}_o$ have unitary norm, it is easy to see that $0 \leq \mathcal{L}_o \leq 2$.
Therefore, $\mathcal{L}_o$ is commensurable to $\mathcal{L}_c$ and the two terms can be safely summed without rescaling them (see Fig.~\ref{fig:cora_cluster} for an example).
$\I_K$ can be interpreted as a (rescaled) clustering matrix $\I_K = \hat{\S}^T\hat{\S}$, where $\hat{\S}$ assigns exactly $N/K$ points to each cluster. 
The value of the Frobenius norm between clustering matrices is not biased by differences in the size of the clusters~\citep{law2017deep} and, thus, can be used to optimize intra-cluster variance.

While traditional SC requires to compute the spectral decomposition for every new sample, here the cluster assignments are computed by a neural network that learns a mapping from the \textit{nodes feature space} to the \textit{clusters assignment space}. 
Since the neural network parameters are independent of the graph size, and since the MP operations in the GNN are localized in the node space and independent from the spectrum of the Laplacian, the proposed clustering approach generalizes to unseen graphs at inference time.
This also gives the opportunity of training our network on small graphs and then use it to cluster larger ones.

%%%%%%%%%%%%%%%%%%%%%%%%%%%%%%%%%%%%%%%%%%%%%%%%%%%
%% POOLING
%%%%%%%%%%%%%%%%%%%%%%%%%%%%%%%%%%%%%%%%%%%%%%%%%%%
\section{Pooling and Graph Coarsening}

\begin{figure}
\centering
      \includegraphics[width=0.85\columnwidth]{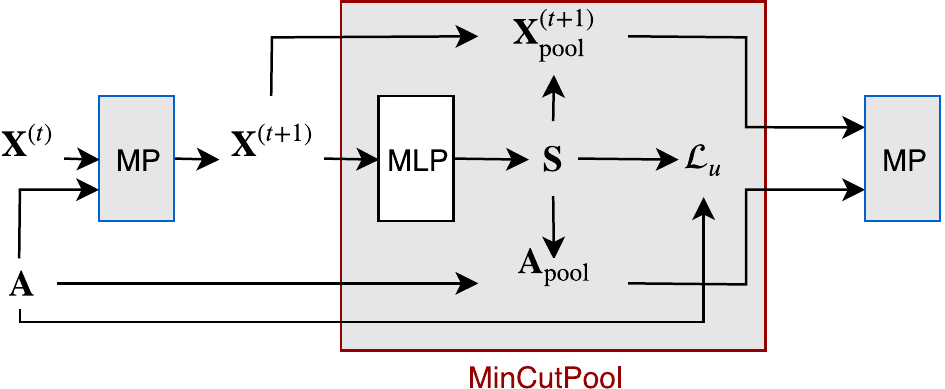}
    \caption{Schema of the \mincutpool{} layer.}
    \label{fig:scheme_mincut}
\end{figure}

The methodology proposed in Section \ref{sec:method} is a general technique that can be used to solve clustering tasks on any data represented by graphs.
In this work, we focus on using it to perform pooling in GNNs and introduce the \mincutpool{} layer, which exploits the cluster assignment matrix $\S$ in \eqref{eq:assigments} to generate a coarsened version of the graph. 
Fig.~\ref{fig:scheme_mincut} depicts a scheme of the \mincutpool{} layer.

The coarsened adjacency matrix and the pooled vertex features are computed, respectively, as
\begin{equation}
    \label{eq:coarsening}
   \A^{pool} = \S^T \tilde \A \S; \;\;\; \X^{pool} = \S^T \X,
\end{equation}
where the entry $x_{i,j}^{pool}$ in $\X^{pool} \in \R^{K \times F}$ is the sum of feature $j$ among the elements in cluster $i$, weighted by the cluster assignment scores.
$\A^{pool} \in \R^{K \times K}$ is a symmetric matrix, whose entries $a^{pool}_{i,i}$ indicate the weighted sum of edges between the nodes in the cluster $i$, while $a^{pool}_{i,j}$ is the weighted sum of edges between cluster $i$ and $j$.
Since $\A^{pool}$ corresponds to the numerator of $\mathcal{L}_c$ in \eqref{eq:coarsening}, the trace maximization yields clusters with many internal connections and weakly connected to each other.
Hence, $\A^{pool}$ will be a diagonal-dominant matrix that describes a graph with self-loops much stronger than any other connection.
Since very strong self-loops hamper the propagation across adjacent nodes in the MP operations following the pooling layer, we compute the new adjacency matrix $\tilde \A^{pool}$ by zeroing the diagonal and applying degree normalization
\begin{equation*}
    %\label{eq:diag_zero}
    \hat \A = \A^{pool} - \I_K \text{diag}(\A^{pool}); \;\;\; \tilde \A^{pool} = \hat \D^{-\frac{1}{2}} \hat \A \hat \D^{-\frac{1}{2}}.
\end{equation*}

Because our GNN-based implementation of SC is fully differentiable, \mincutpool{} layers can be used to build deep GNNs that hierarchically coarsen the graph representation. 
The parameters of each \mincutpool{} layer can then be learned end-to-end, by jointly optimizing $\mathcal{L}_u$ along with any supervised loss for a particular downstream task.
Contrarily to SC methods that search for feasible solutions only within the space of orthogonal matrices, $\mathcal{L}_o$ only introduces a soft constraint that can be partially violated during the learning procedure.
This allows the GNN to find the best trade-off between $\mathcal{L}_u$ and the supervised loss, and makes it possible to handle graphs with intrinsically imbalanced clusters.
Since $\mathcal{L}_c$ is non-convex, the violation of the orthogonality constraint could compromise the convergence to the global optimum of the \mincut{} objective.
However, we note that: 
\begin{enumerate}
    \item since \mincutpool{} computes the cluster assignments from node features that become similar due to MP operations, clusters are likely to contain nodes that are both strongly connected and with similar features, reducing the risk of finding a degenerate solution;
    \item the degenerate minima of $\mathcal{L}_c$ lead to discarding most of the information from the input graph and, therefore, optimizing the task-specific loss encourages the GNN to avoid them;
    \item since the \mincut{} objective acts mostly as a regularization term, a solution that is sub-optimal for \eqref{eq:ncut_relax} could instead be preferable for the supervised downstream task;
\end{enumerate}
For these reasons, we show in Section \ref{sec:experiments} that \mincutpool{} never yields degenerate solutions in practice, but consistently achieves good performance on a variety of tasks. 

\subsection{Computational Complexity}
The space complexity of the \mincutpool{} layer is $\mathcal{O}(NK)$, as it depends on the dimension of the cluster assignment matrix $\S \in \R^{N \times K}$.
The computational complexity is dominated by the numerator in the term $\mathcal{L}_c$, and is $\mathcal{O}(N^2K + NK^2) = \mathcal{O}(NK(K+N))$.
Since $\tilde \A$ is usually sparse, we can exploit operations for sparse tensors and reduce the complexity of the first matrix multiplication to $\mathcal{O}(EK)$, where $E$ is the number of non-zero edges in $\tilde \A$.
Since the sparse multiplication yields a dense matrix, the second multiplication still costs $\mathcal{O}(NK^2)$ and the total cost is $\mathcal{O}(K(E + NK))$.

\subsection{Related Work on Pooling in GNNs}
\label{sec:related}
In this section, we summarize some related works on graph pooling in GNNs covering the two main families of methods: model-free and model-based.

\textbf{Model-Free Pooling\;} These methods pre-compute the coarsened graphs based on the topology ($\A$) but do not take explicitly into consideration the original node features ($\X$), or the intermediate node representation produced by the GNN.
During the forward pass of the GNN, the node features are aggregated with simple procedures and matched to the pre-computed graph structures. 
These methods are less flexible and cannot adapt to different tasks, but provide a stronger inductive bias that can help to avoid degenerate solutions (e.g., coarsened graphs collapsing in a single node).

One of the most popular model-free method is Graclus \citep{dhillon2007weighted} (adopted in \cite{defferrard2016convolutional,fey2018splinecnn,monti2017geometric}), which implements an equivalent formulation of SC based on the less expensive kernel k-means algorithm.
At each pooling level, Graclus identifies pairs of maximally similar nodes to be clustered together into a new vertex. 
During each forward pass of the GNN, max pooling is used to determine the node attribute to keep from each pair.
Fake vertices are added so that the number of nodes can be halved each time, although this injects noisy information in the graph.

\textit{Node decimation pooling} (NDP) is a method originally proposed in graph signal processing literature~\citep{shuman2016multiscale}, which has been adapted to perform pooling in GNNs~\citep{simonovsky2017dynamic, bianchi2019hierarchical}.
The nodes are partitioned in two sets, according to the signs of the Laplacian eigenvector associated with the largest eigenvalue.
One of the two sets is dropped, reducing the number of nodes by approximately half. Kron reduction is then used to generate the coarsened graphs by connecting the remaining nodes.

\textbf{Model-Based Pooling\;} These approaches learn how to a coarsen graphs through learnable functions, which take as input the nodes features $\X$ and are parametrized by weights optimized on the task and data at hand.
While being more flexible, model-based approaches are mostly based on heuristics and, as shown in Sec.~\ref{sec:experiments}, tend to fail in several cases, leading to instability in the training process.

\textit{DiffPool}~\citep{ying2018hierarchical} is a pooling method that uses a MP layer to compute a clustering of the input graphs. DiffPool was one of the first attempts at learning a pooling operator end-to-end, and is regularized by minimizing the entropy of the cluster assignment along with a link prediction loss.

The approach dubbed \textit{Top-$K$} pooling~\citep{graphunet, lee2019selfattention}, learns a projection vector that is applied to each node feature to obtain a score. The nodes with the $K$ highest scores are retained, while the others are dropped. 
Since the top-$K$ selection is not differentiable, the scores are also used as a gate (or attention) for the node features, letting the projection vector be trained with backpropagation.
Top-$K$ is generally less effective than DiffPool in benchmarks but is significantly more memory efficient as it avoids generating dense cluster assignments.

We conduct an in-depth comparative analysis of \mincutpool{}, DiffPool, and Top-$K$ in Section \ref{sec:experiments}, showing some significant drawbacks of the two previous learnable approaches w.r.t.\ to our method, and highlighting how the inductive bias inherited by SC leads to significant improvements in performance in \mincutpool{}. 

%%%%%%%%%%%%%%%%%%%%%%%%%%%%%%%%%%%%%%%%%%%%%%%%%%%
%% EXPERIMENTS
%%%%%%%%%%%%%%%%%%%%%%%%%%%%%%%%%%%%%%%%%%%%%%%%%%%
\section{Experiments}
\label{sec:experiments}
We consider both supervised and unsupervised tasks to compare \mincutpool{} with traditional SC and with other GNN pooling strategies.
The Appendix provides further details on the experiments and a schematic depiction of the architectures used in each task.
In addition, the Appendix reports an additional experiment on supervised graph regression.
The implementation of \mincutpool{} is available both in Spektral\footnote{\url{https://graphneural.network/layers/pooling/\#mincutpool}} and Pytorch Geometric.\footnote{\url{https://pytorch-geometric.readthedocs.io/en/latest/modules/nn.html\#torch_geometric.nn.dense.mincut_pool.dense_mincut_pool}}

\subsection{Clustering the Graph Nodes}
\begin{figure}
    \centering
    \subfigure[SC]{
        \includegraphics[width=0.3\columnwidth]{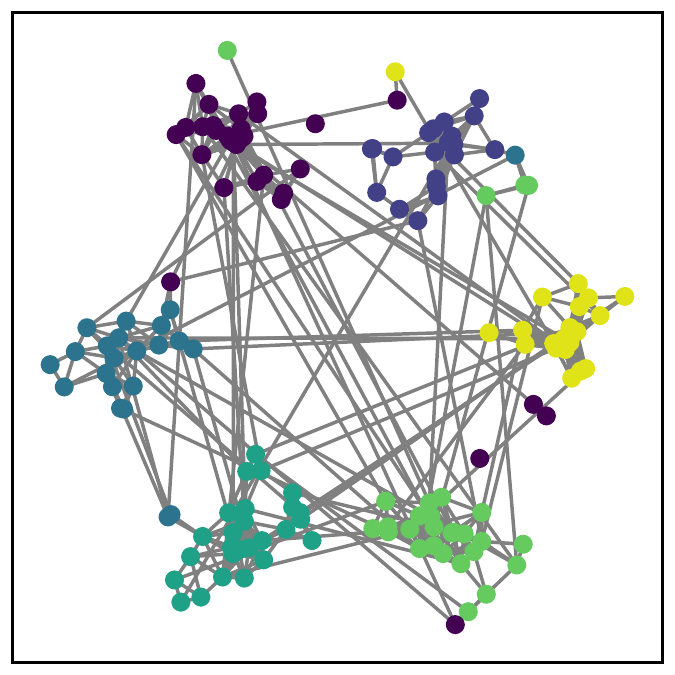}
    }%\hspace{-0.5em}%
    ~
    \subfigure[DiffPool]{
        \includegraphics[width=0.3\columnwidth]{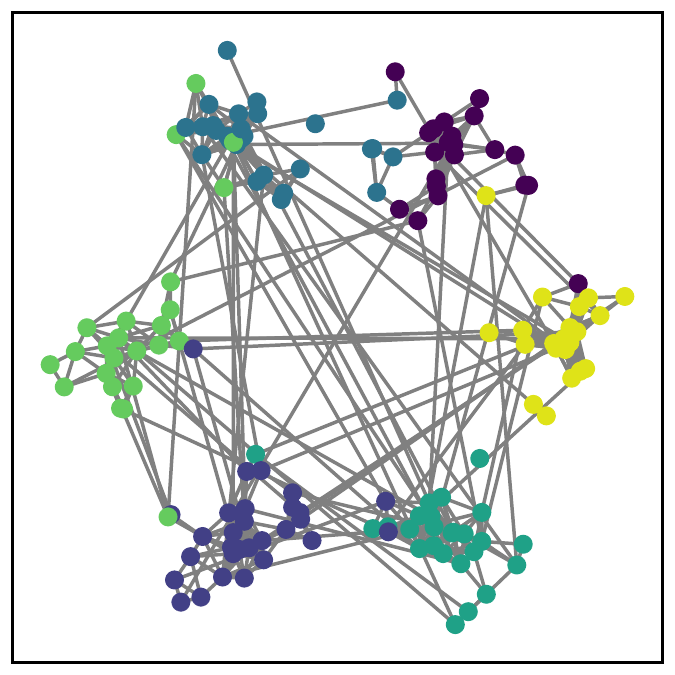}
    }%\hspace{-0.5em}%
    ~
    \subfigure[\mincutpool{}]{
        \includegraphics[width=0.3\columnwidth]{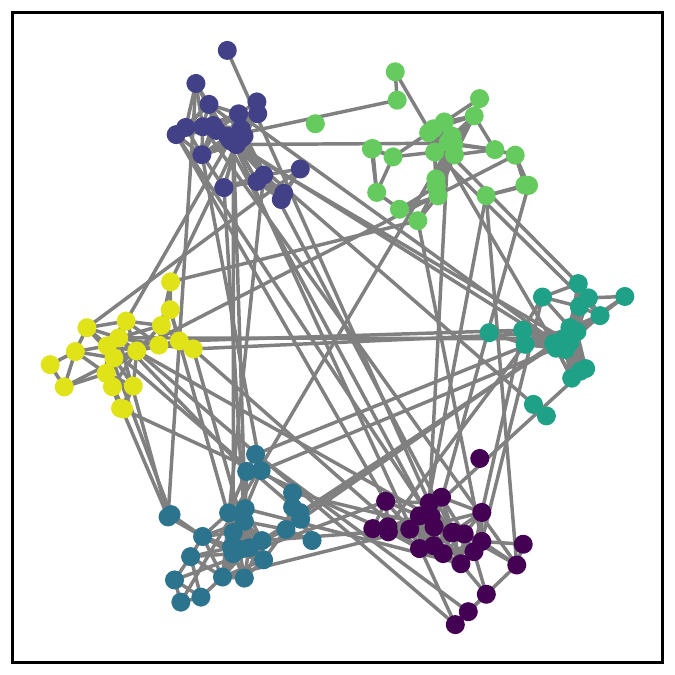}
    }
    
    \subfigure[SC]{
        \includegraphics[width=0.3\columnwidth]{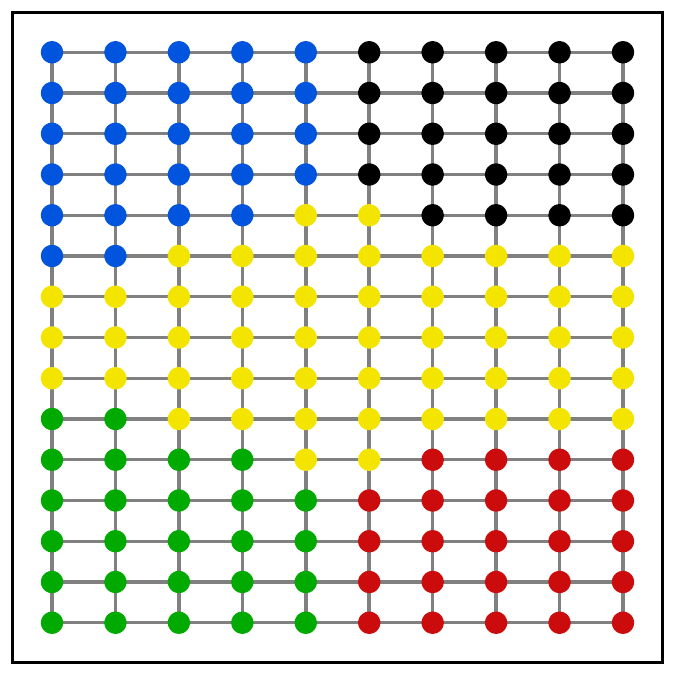}
    }%\hspace{-.5em}%
    ~
    \subfigure[DiffPool]{
        \includegraphics[width=0.3\columnwidth]{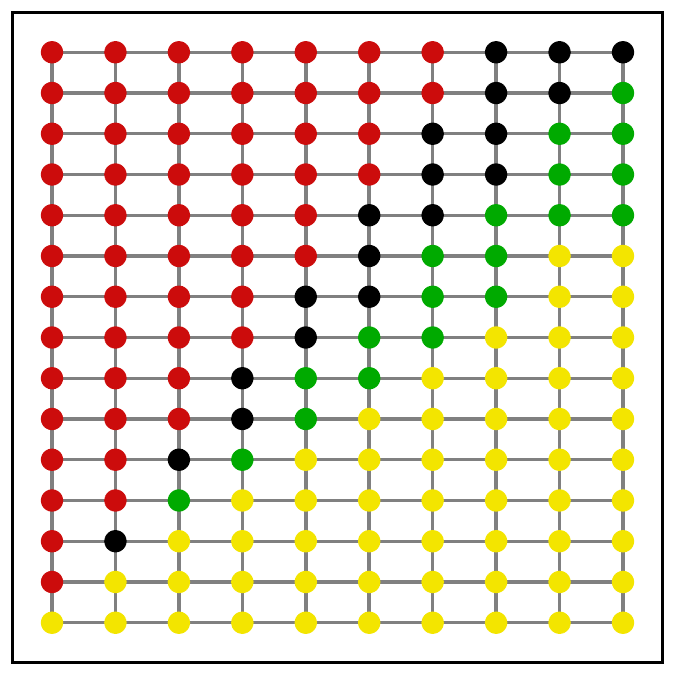}
    }%\hspace{-.5em}%
    ~
    \subfigure[\mincutpool{}]{
        \includegraphics[width=0.3\columnwidth]{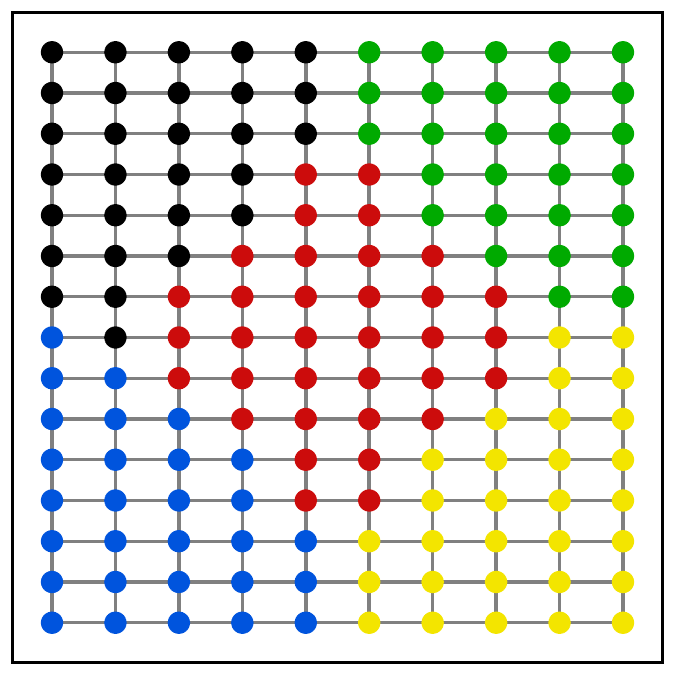}
    }
    \caption{Node clustering on a community network ($K$=6) and on a grid graph ($K$=5).}
    \label{fig:synth_cluster}
\end{figure}

In this experiment, we evaluate the quality of the assignments $\S$ found by our method on two unsupervised tasks.
We implement a one-layer GNN followed by a single-layer MLP to compute $\S$, and train the overall architecture by minimizing $\mathcal{L}_u$.
For comparison, we configure a similar GNN architecture based on DiffPool where we optimize the auxiliary DiffPool losses (see Sec.~\ref{sec:related}) without any additional supervised loss. We also consider the clusters found by SC, which, unlike our approach, are based on the spectrum of the graph.
In the results, our approach is always indicated as \mincutpool{} for simplicity, although this experiment only focuses on the clustering results and does not involve the coarsening step. A similar consideration holds for DiffPool.

\begin{figure}[!ht]
    \centering
    \subfigure[DiffPool]{
        \includegraphics[width=.8\columnwidth]{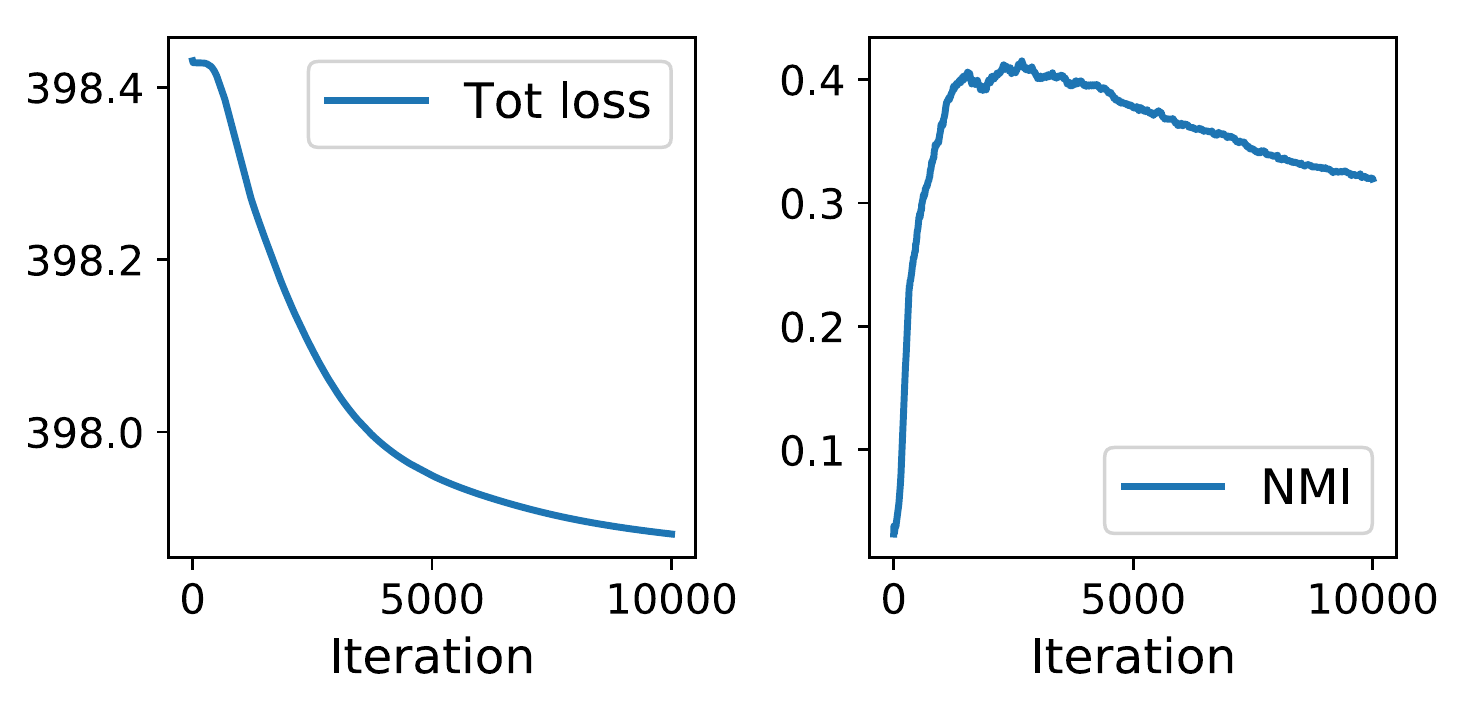}
    }\hspace{-1.35em}%
    \subfigure[\mincutpool{}]{
        \includegraphics[width=.8\columnwidth]{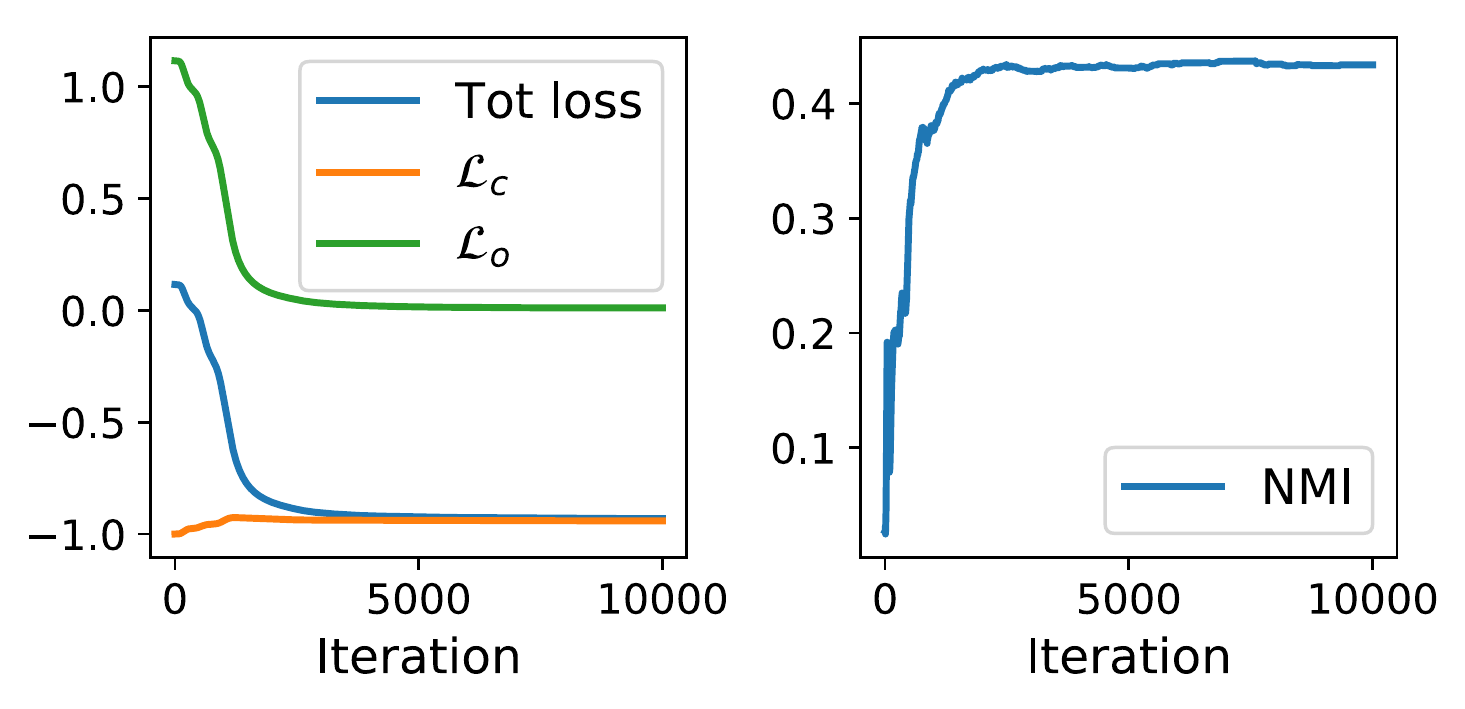}
    }\hspace{-0.5em}%
    \caption{Unsupervised losses and NMI of DiffPool and \mincutpool{} on Cora.}
    \label{fig:cora_cluster}
\end{figure}

\textbf{Clustering on Synthetic Networks\;} We consider two simple graphs: the first is a network with 6 communities and the second is a regular grid. 
The adjacency matrix $\A$ is binary and the features $\X$ are the 2-D node coordinates.
Fig.~\ref{fig:synth_cluster} depicts the node partitions generated by SC (a, d), DiffPool (b, e), and \mincutpool{} (c, f).
\mincutpool{} generates very accurate and balanced partitions, demonstrating that the cluster assignment matrix $\S$ is well-formed. In particular, we note that the inductive bias carried by the node features in \mincutpool{} leads to a different clustering than SC but results in an overall better performance (c.f.\ following discussion and Figure \ref{fig:cora_cluster}).
On the other hand, DiffPool assigns some nodes to the wrong community in the first example and produces an unbalanced partition of the grid.

\begin{figure}[!t]
    \centering
    \subfigure[Original image]{
        \includegraphics[width=0.49\columnwidth]{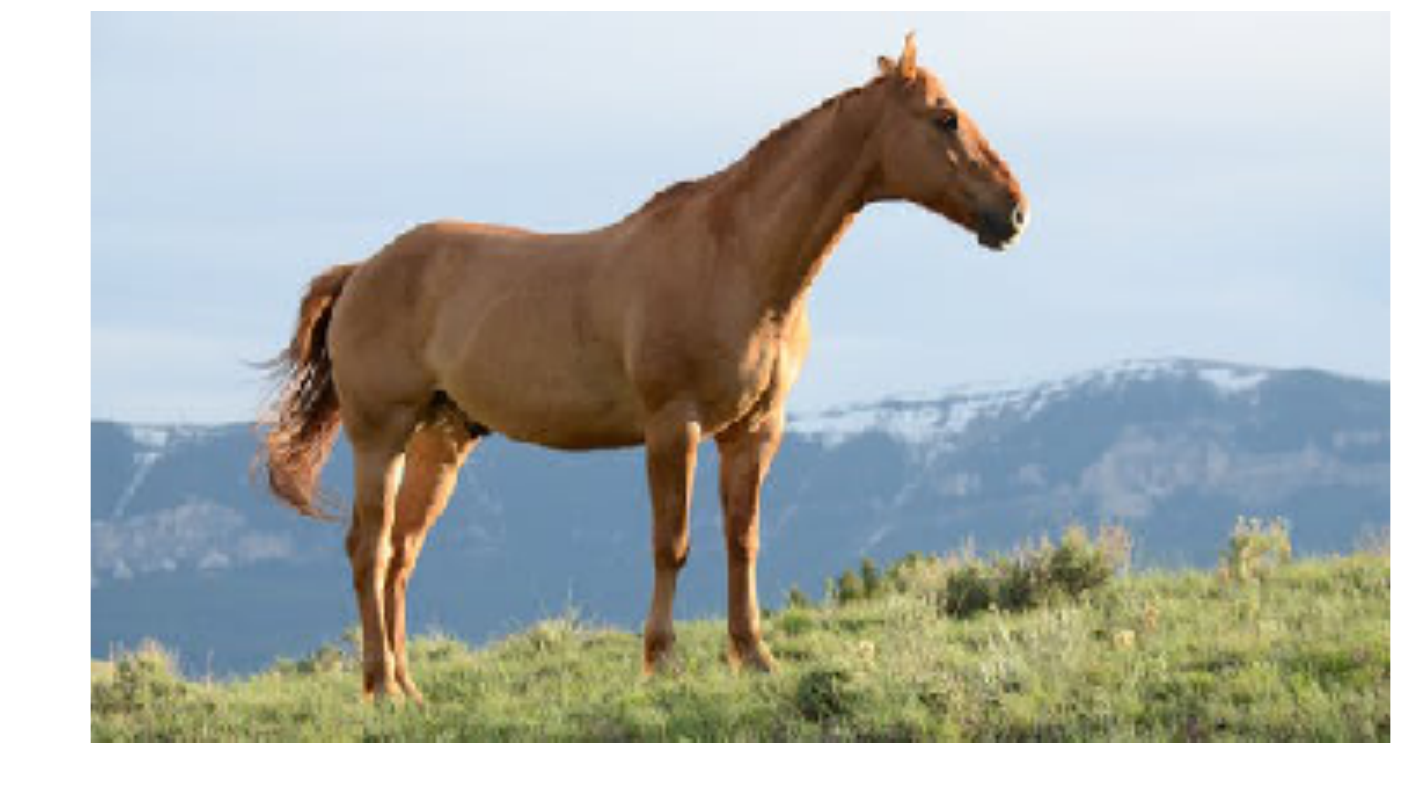}
    }\hspace{-1em}%
    ~
    \subfigure[Oversegmentation]{
        \includegraphics[width=0.49\columnwidth]{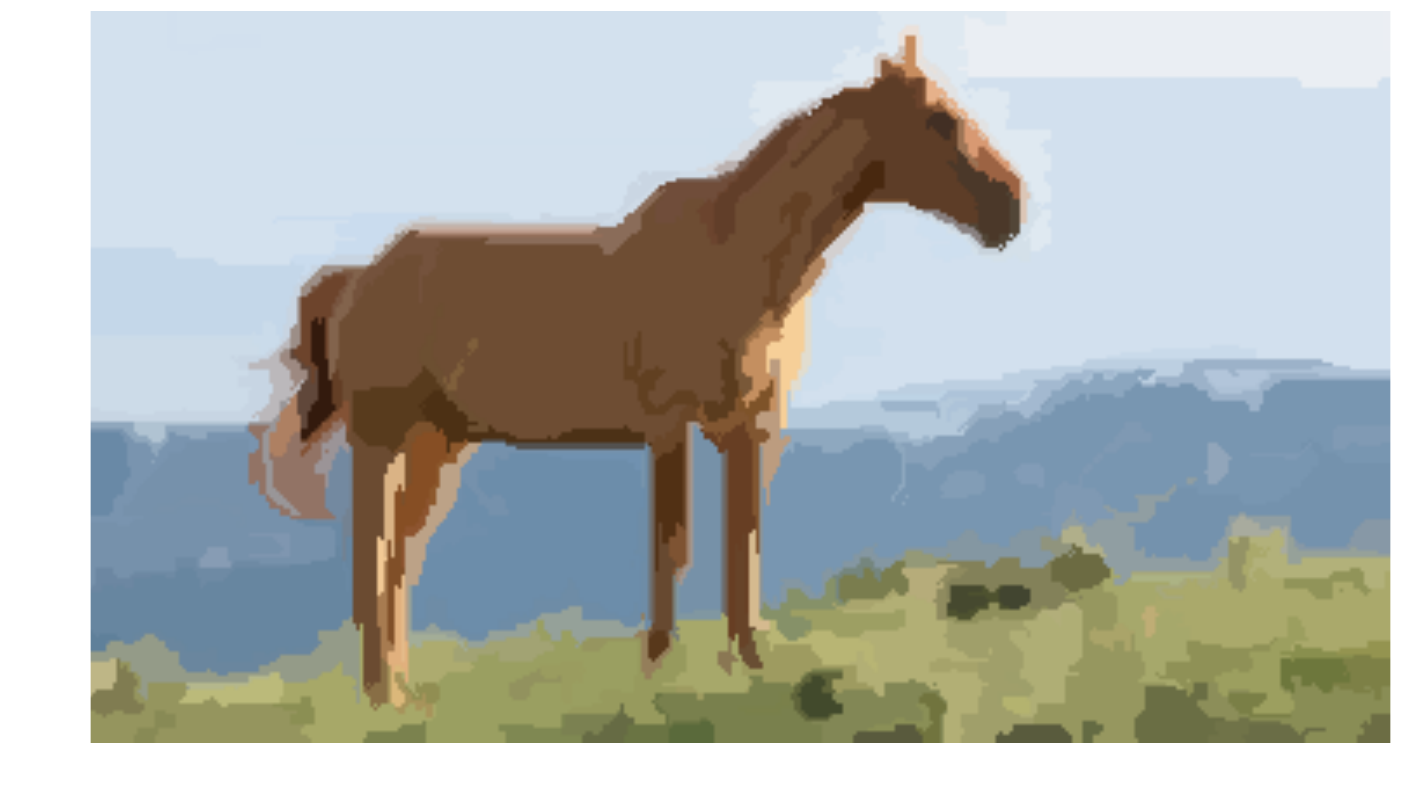}
    }\hspace{-1em}%
    
    \subfigure[Region Adjacency Graph]{
        \includegraphics[width=0.49\columnwidth]{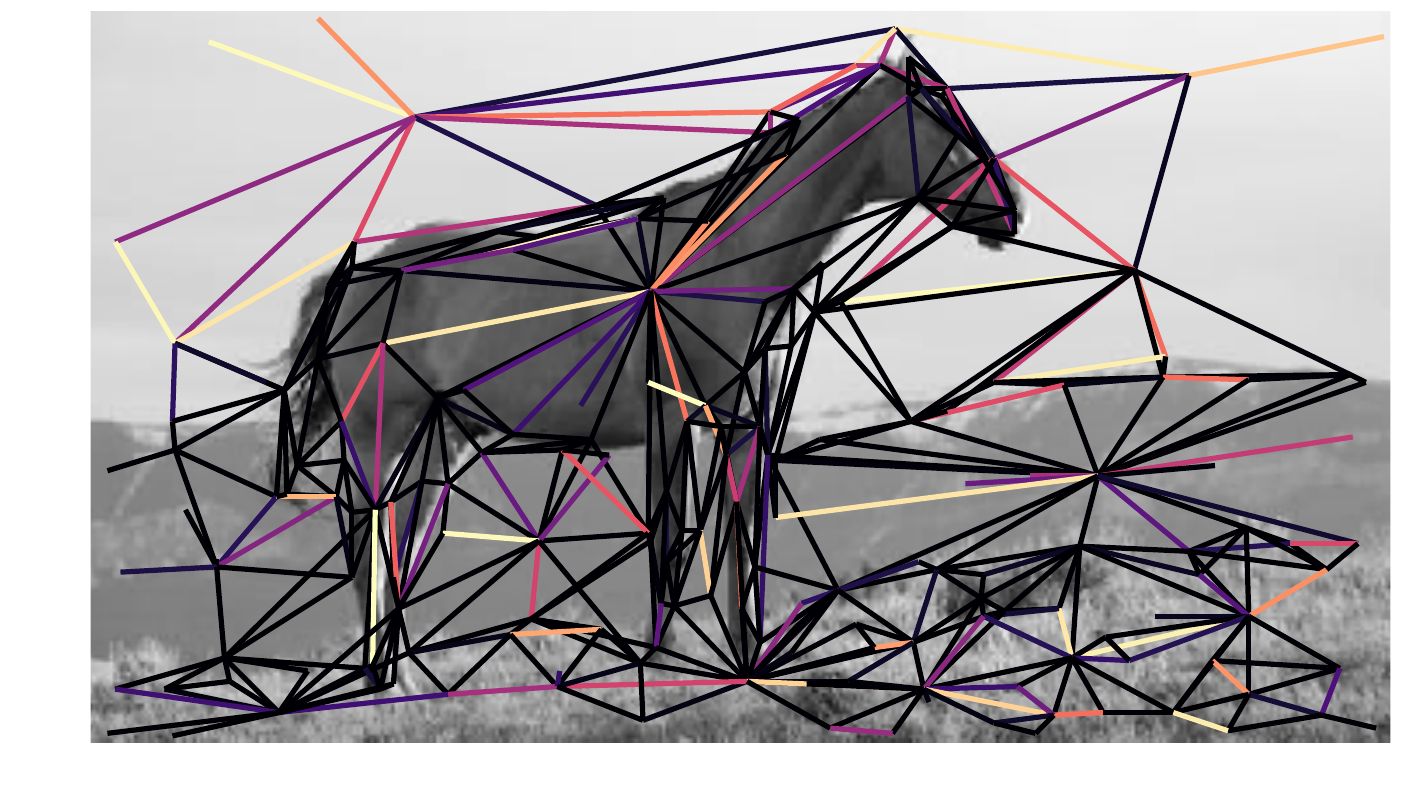}
    }\hspace{-1em}%
    ~
    \subfigure[SC]{
        \includegraphics[width=0.49\columnwidth]{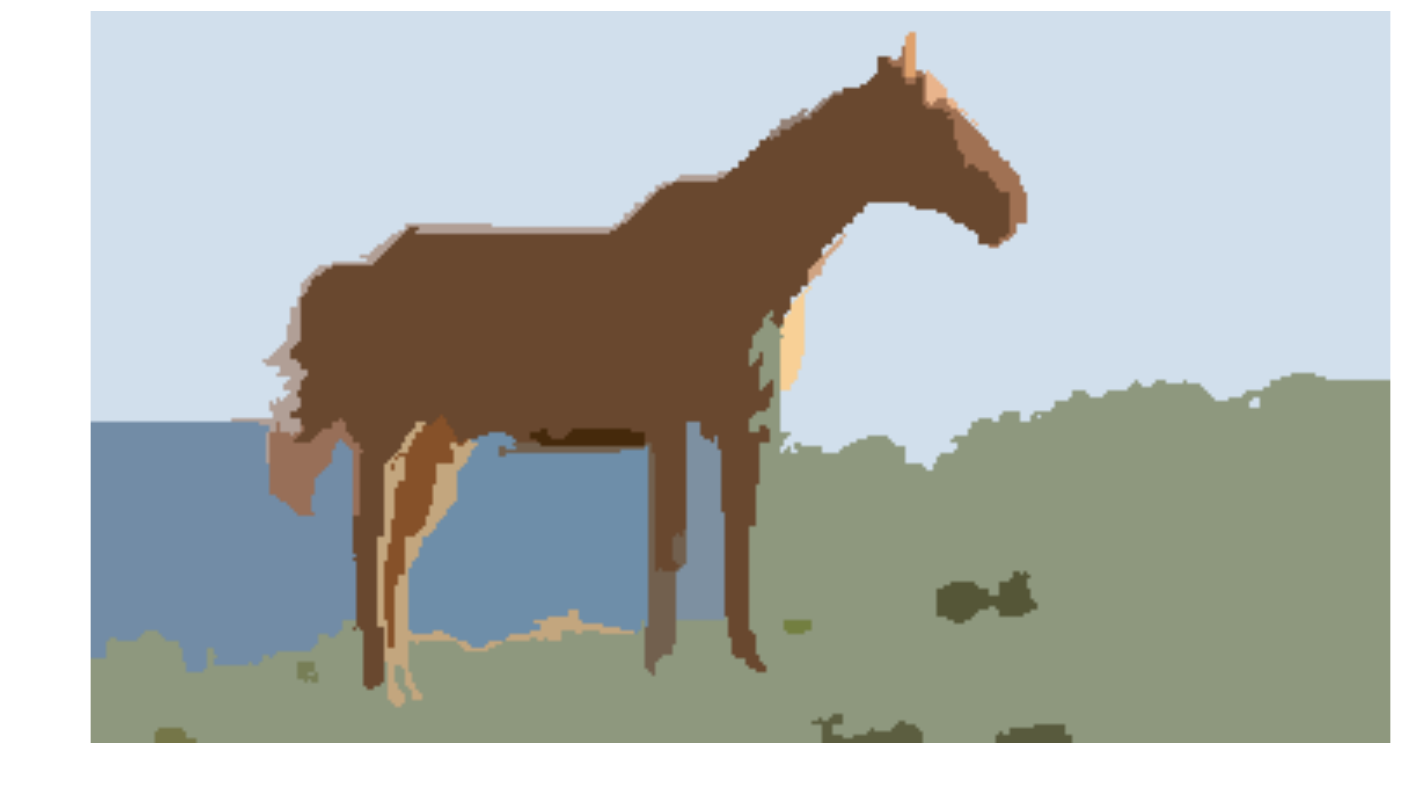}
    }\hspace{-1em}%
    
    \subfigure[DiffPool ($K=4$)]{
        \includegraphics[width=0.49\columnwidth]{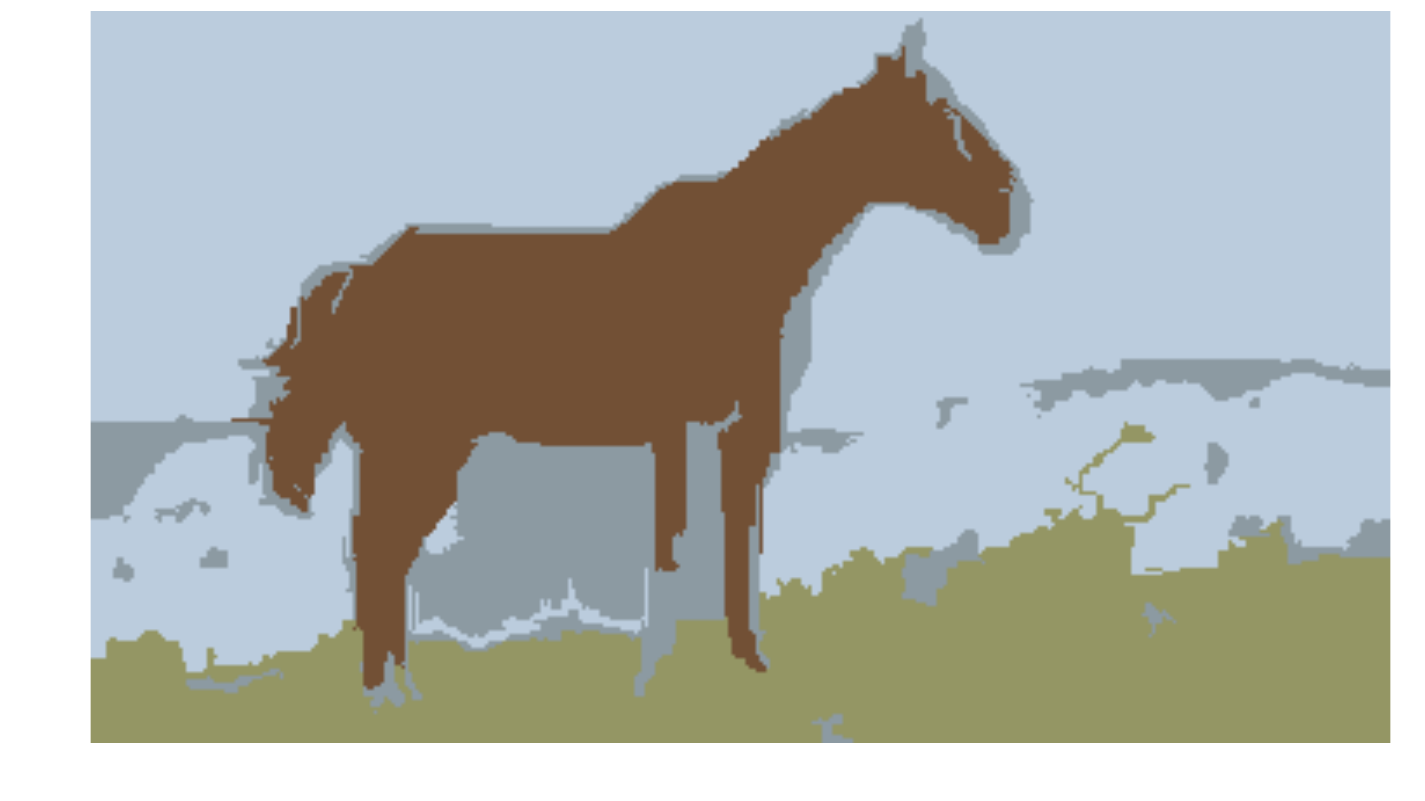}
    }\hspace{-1em}%
    ~
    \subfigure[\mincutpool{} ($K=4$)]{
        \includegraphics[width=0.49\columnwidth]{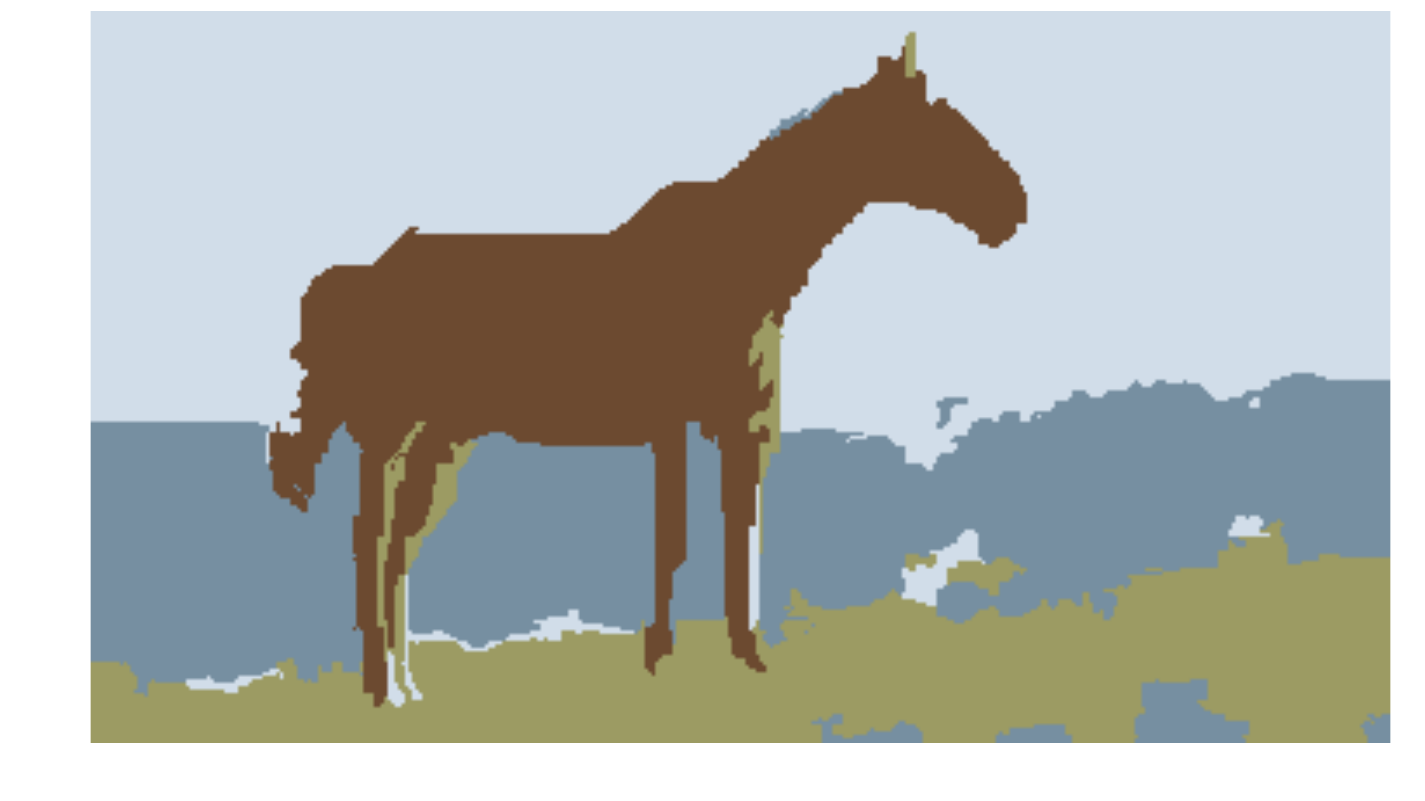}
    }\hspace{-1em}%
    
    \caption{Image segmentation by clustering the nodes of the Region Adjacency Graph.}
    \label{fig:segmentation}
\end{figure}

\textbf{Image Segmentation\;} Given an image, we build a region adjacency graph~\citep{tremeau2000regions} using as nodes the regions generated by an over-segmentation procedure~\citep{felzenszwalb2004efficient}.
The SC technique used in this example is the recursive normalized cut~\citep{shi2000normalized}, which recursively clusters the nodes until convergence.
For \mincutpool{} and DiffPool, node features consist of the average and total colour in each over-segmented region. We set the number of desired clusters to $K=4$.
The results in Fig.~\ref{fig:segmentation} show that \mincutpool{} yields a precise and intuitive segmentation. On the other hand, SC and DiffPool aggregate wrong regions and, also, SC finds too many segments.

\textbf{Clustering on Citation Networks\;} We cluster the nodes of three citation networks: Cora, Citeseer, and Pubmed. 
The nodes are documents represented by sparse bag-of-words feature vectors and the binary undirected edges indicate citation links between documents.
Each node is labelled with the document class, which we use as ground truth for the clusters. 
To evaluate the partitions generated by each method, we check the agreement between the cluster assignments and the true labels.
Tab.~\ref{tab:clust_cit} reports the Completeness Score $\text{CS}(\tilde{\y}, \y) = 1 - \frac{H(\tilde{\y}|\y)}{H(\tilde{\y})}$ and Normalized Mutual Information $\text{NMI}(\tilde{\y}, \y) = \frac{H(\tilde{\y}) - H(\tilde{\y}|\y)}{\sqrt{H(\tilde{\y}) - H(\y)}}$, where $H(\cdot)$ is the entropy.

Once again, our GNN architecture achieves a higher NMI score than SC, which does not account explicitly for the node features when generating the clusters.
\mincutpool{} outperforms also DiffPool, since the minimization of the unsupervised loss in DiffPool fails to converge to a good solution.
A pathological behaviour in DiffPool is revealed by Fig.~\ref{fig:cora_cluster}, which depicts the evolution of the NMI scores as the unsupervised losses in DiffPool and \mincutpool{} are minimized in training (note how the NMI of DiffPool eventually decreases).
From Fig.\ \ref{fig:cora_cluster}, it is also possible to see the interaction between the \mincut{} loss and the orthogonality loss in our approach. 
In particular, the \mincut{} loss does not converge to its minimum ($\mathcal{L}_c = -1$), corresponding to one of the degenerate solutions discussed in Sec.~\ref{sec:method}. 
Instead, \mincutpool{} learns the optimal trade-off between $\mathcal{L}_c$ and $\mathcal{L}_o$ and achieves a better and more stable clustering performance than DiffPool. 

\begin{table*}
\footnotesize
\centering
\caption{NMI and CS obtained by clustering the nodes on citation networks over 10 different runs. The number of clusters $K$ is equal to the number of node classes.} 
\label{tab:clust_cit}
\begin{tabular}{lccccccc}
\cmidrule[1.5pt]{1-8}
\textbf{Dataset} & $\boldsymbol{K}$ & \multicolumn{2}{c}{\textbf{Spectral clustering}} & \multicolumn{2}{c}{\textbf{DiffPool}} & \multicolumn{2}{c}{\textbf{\mincutpool{}}} \\
\cmidrule[.5pt]{1-8}
% Values for spectral clustering computed over 256 runs
% Values for Cora/Citeseer/Pubmed computed ovevr 10 runs
% Hparams for Mincut/DiffPool: 16 channels, no MLP, lr=0.0005, ELU activation, 10000 iterations
            &   & NMI  & CS & NMI & CS & NMI & CS \\
Cora        & 7 & 0.025 \t{$\pm$ 0.014} & 0.126 \t{$\pm$ 0.042} & 0.315 \t{$\pm$ 0.005} & 0.309 \t{$\pm$ 0.005} & \textbf{0.404} \t{$\pm$ 0.018} & \textbf{0.392} \t{$\pm$ 0.018} \\
Citeseer    & 6 & 0.014 \t{$\pm$ 0.003} & 0.033 \t{$\pm$ 0.000} & 0.139 \t{$\pm$ 0.016} & 0.153 \t{$\pm$ 0.020} & \textbf{0.287} \t{$\pm$ 0.047} & \textbf{0.283} \t{$\pm$ 0.046} \\
Pubmed      & 3 & 0.182 \t{$\pm$ 0.000} & \textbf{0.261} \t{$\pm$ 0.000} & 0.079 \t{$\pm$ 0.001} & 0.085 \t{$\pm$ 0.001} & \textbf{0.200} \t{$\pm$ 0.020} & 0.197 \t{$\pm$ 0.019} \\
\cmidrule[1.5pt]{1-8}
\end{tabular}
\end{table*}

\subsection{GNN Autoencoder}

\begin{figure}[!t]
    \centering
    \subfigure[Original]{
        \includegraphics[width=0.32\columnwidth]{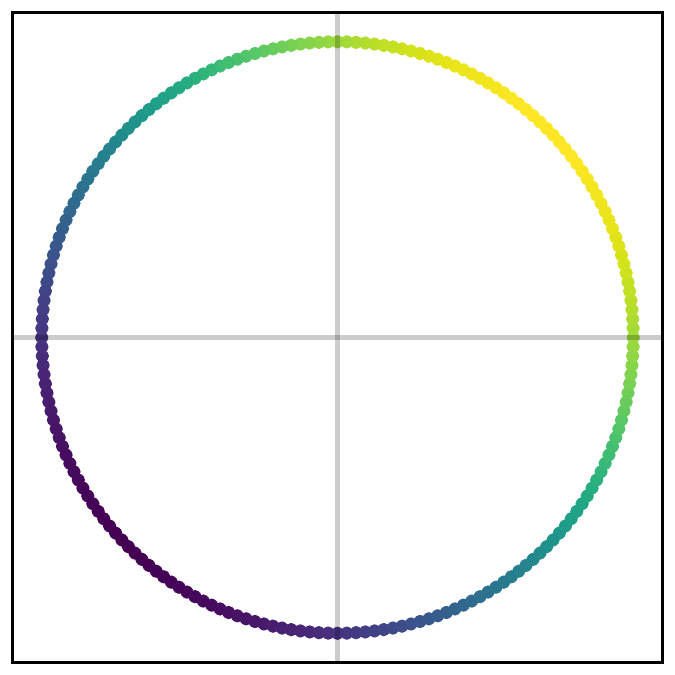}
    }\hspace{-0.5em}%
    ~
    \subfigure[Top-$K$]{
        \includegraphics[width=0.32\columnwidth]{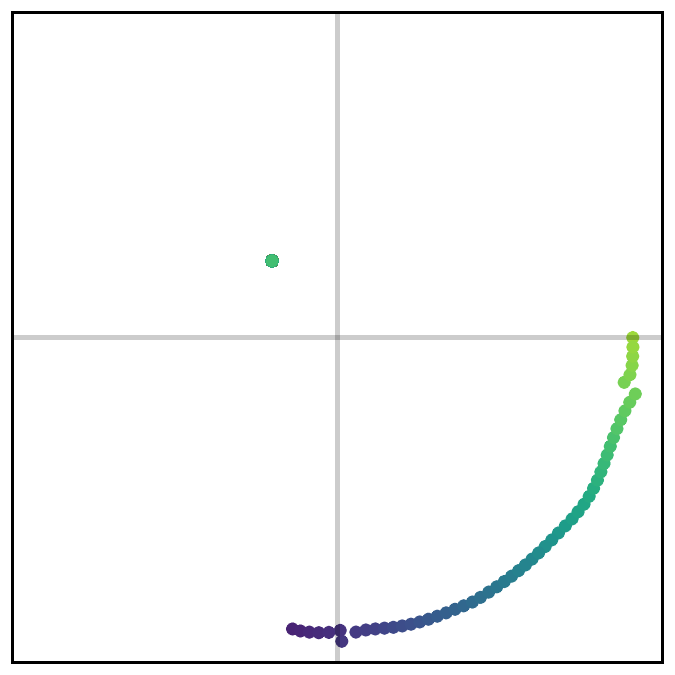}
    }\hspace{-0.5em}%
    ~
    \subfigure[DiffPool]{
        \includegraphics[width=0.32\columnwidth]{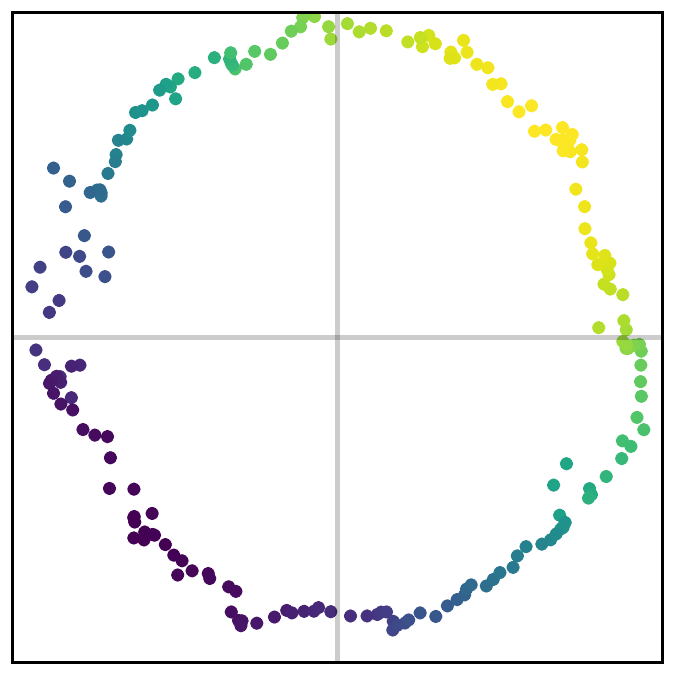}
    }\hspace{-0.5em}%
    ~
    \subfigure[\mincutpool{}]{
        \includegraphics[width=0.32\columnwidth]{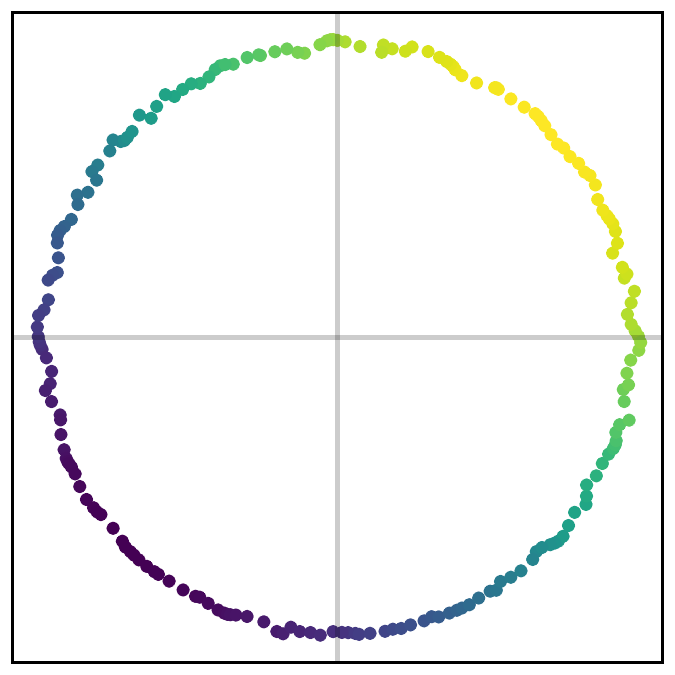}
    }
    \caption{AE reconstruction of a ring graph}
    \label{fig:AE_reconstr_circ}
\end{figure}

\begin{figure}[!t]
    \centering
    \subfigure[Original]{
        \includegraphics[width=0.32\columnwidth]{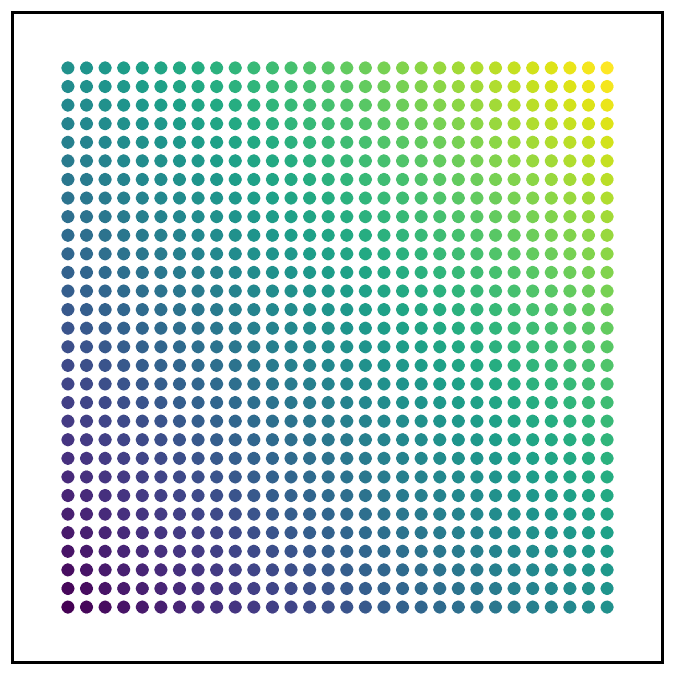}
    }\hspace{-0.5em}%
    ~
    \subfigure[Top-$K$]{
        \includegraphics[width=0.32\columnwidth]{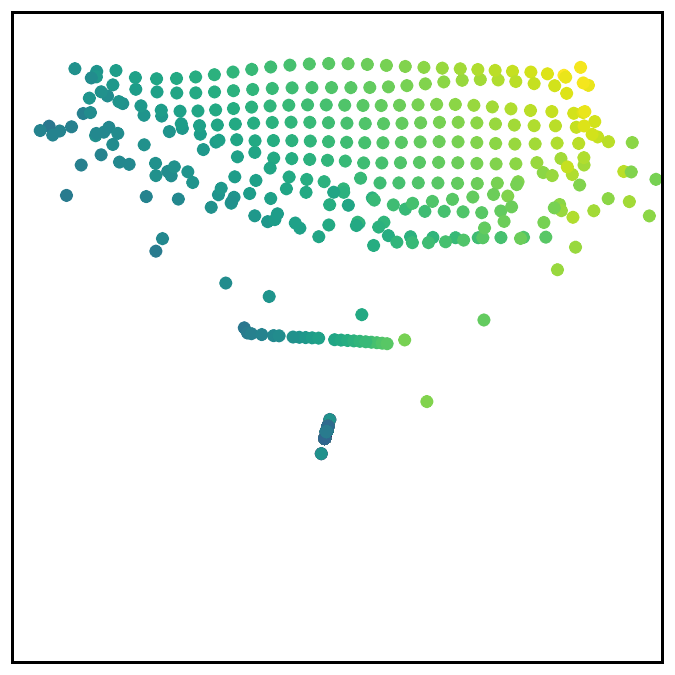}
    }\hspace{-0.5em}%
    ~
    \subfigure[DiffPool]{
        \includegraphics[width=0.32\columnwidth]{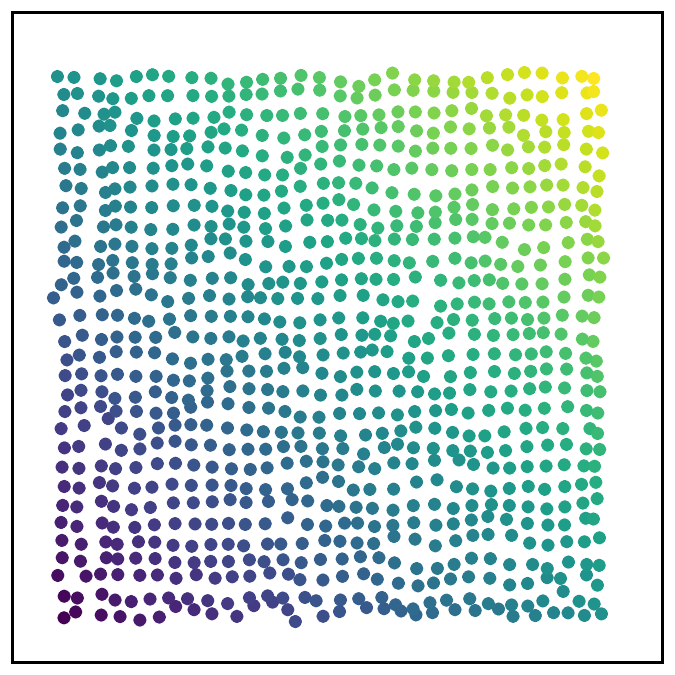}
    }\hspace{-0.5em}%
    ~
    \subfigure[\mincutpool{}]{
        \includegraphics[width=0.32\columnwidth]{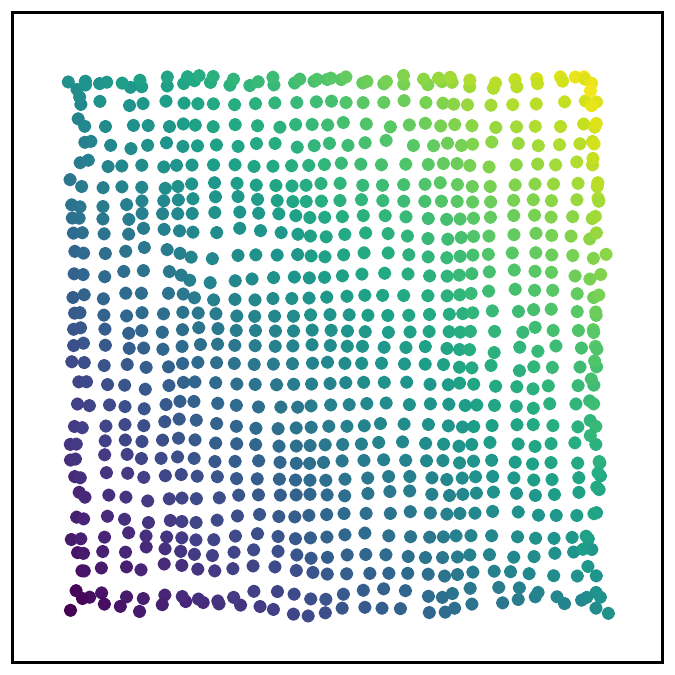}
    }
    \caption{AE reconstruction of a grid graph}
    \label{fig:AE_reconstr_grid}
\end{figure}
To quantify the amount of information retained by different pooling layers, we train an autoencoder (AE) to reconstruct an input graph signal $\X$ from its pooled version. We compare \mincutpool{}, DiffPool, and Top-$K$.
The AE is trained by minimizing the mean squared error between the original and the reconstructed graph features, $\|\X - \X^\text{rec}\|^2$.
All the pooling operations keep $25\%$ of the original nodes.

To upscale the coarsened graph back to its original size, in \mincutpool{} we transpose the pooling operations:
\begin{equation}
    \label{eq:unpool}
    \X^\text{rec} = \S \X^\text{pool}; \;\; \A^\text{rec} = \S \A^\text{pool} \S^T.
\end{equation}
A similar operation is performed for DiffPool. For Top-$K$, we use the un-pooling operation from the original paper~\cite{graphunet}.

Fig.~\ref{fig:AE_reconstr_circ} and \ref{fig:AE_reconstr_grid} report the original graph signal $\X$ (the node features are the 2-D coordinates of the nodes) and the reconstruction $\X^\text{rec}$ obtained by the different pooling methods, for a ring and a regular grid graph.
The reconstruction produced by DiffPool is worse for the ring graph, but is good for the grid graph, while \mincutpool{} yields almost perfect results in both cases.
On the other hand, Top-$K$ fails to generate a coarsened graph that maintains enough information to reconstruct the original graph.

This experiment highlights a major issue in Top-$K$ pooling, which retains the nodes associated with the highest $K$ values of a score vector $\mathbf{s}$, computed by projecting the node features onto a trainable vector $\mathbf{p}$: $\mathbf{s} = \X \mathbf{p}$.
Nodes that are connected on the graph usually share similar features, and their similarity further increases after the MP operations, which combine the features of neighbouring nodes.
Retaining the nodes associated with the highest $K$ scores in $\mathbf{s}$ corresponds to keeping those nodes that are alike and highly connected, as it can be seen in Fig.~\ref{fig:AE_reconstr_circ}-\ref{fig:AE_reconstr_grid}.
Therefore, Top-$K$ drops entire portions of the graph, making it impossible to recover the discarded information.
This explains why Top-$K$ fails to recover the original graph signal when used as bottleneck for the AE, and motivates the lower performance of Top-$K$ in graph classification (Sec.\ \ref{sec:node_class}).

\begin{table*}[!h]
\footnotesize
\centering
\caption{Graph classification accuracy. Significantly better results ($p < 0.05$) are in bold.} 
\label{tab:graph_class}
\begin{tabular}{lccccccccc}
\cmidrule[1.5pt]{1-10}
\textbf{Dataset} & \textbf{WL} & \textbf{Dense} & \textbf{No-pool} & \textbf{Graclus} & \textbf{NDP} & \textbf{DiffPool} & \textbf{Top-$K$} & \textbf{SAGpool} & \textbf{\mincutpool{}} \\
\cmidrule[.5pt]{1-10}
Bench-easy    & 92.6 & 29.3\t{$\pm$0.3} & 98.5\t{$\pm$0.3} & 97.5\t{$\pm$0.5} & 97.9\t{$\pm$0.5} & 98.6\t{$\pm$0.4} & 82.4\t{$\pm$8.9} & 84.2\t{$\pm$2.3} & \textbf{99.0\t{$\pm$0.0}} \\
Bench-hard    & 60.0 & 29.4\t{$\pm$0.3} & 67.6\t{$\pm$2.8} & 69.0\t{$\pm$1.5} & \textbf{72.6\t{$\pm$0.9}} & 69.9\t{$\pm$1.9} & 42.7\t{$\pm$15.2} & 37.7\t{$\pm$14.5} & \textbf{73.8\t{$\pm$1.9}}\\  \midrule
Mutagenicity  & \textbf{81.7}\t{$\pm$1.1} & 68.4\t{$\pm$0.3} & 78.0\t{$\pm$1.3} & 74.4\t{$\pm$1.8} & 77.8\t{$\pm$2.3} & 77.6\t{$\pm$2.7} & 71.9\t{$\pm$3.7} & 72.4\t{$\pm$2.4} & 79.9\t{$\pm$2.1} \\  
Proteins      & 71.2\t{$\pm$2.6} & 68.7\t{$\pm$3.3} & 72.6\t{$\pm$4.8} & 68.6\t{$\pm$4.6} & 73.3\t{$\pm$3.7} & 72.7\t{$\pm$3.8} & 69.6\t{$\pm$3.5} & 70.5\t{$\pm$2.6} & \textbf{76.5\t{$\pm$2.6}} \\  
DD            & 78.6\t{$\pm$2.7} & 70.6\t{$\pm$5.2} & 76.8\t{$\pm$1.5} & 70.5\t{$\pm$4.8} & 72.0\t{$\pm$3.1} & \textbf{79.3\t{$\pm$2.4}} & 69.4\t{$\pm$7.8} & 71.5\t{$\pm$4.5} & \textbf{80.8\t{$\pm$2.3}} \\  
COLLAB        & 74.8\t{$\pm$1.3} & 79.3\t{$\pm$1.6} & \textbf{82.1\t{$\pm$1.8}} & 77.1\t{$\pm$2.1} & 79.1\t{$\pm$1.5} & 81.8\t{$\pm$1.4} & 79.3\t{$\pm$1.8} & 79.2\t{$\pm$2.0} & \textbf{83.4\t{$\pm$1.7}} \\  
Reddit-Binary & 68.2\t{$\pm$1.7} & 48.5\t{$\pm$2.6} & 80.3\t{$\pm$2.6} & 79.2\t{$\pm$0.4} & 84.3\t{$\pm$2.4} & 86.8\t{$\pm$2.1} & 74.7\t{$\pm$4.5} & 73.9\t{$\pm$5.1}  & \textbf{91.4\t{$\pm$1.5}} \\  
\cmidrule[1.5pt]{1-10}
\end{tabular}
\end{table*}

\subsection{Supervised Graph Classification}
\label{sec:node_class}

In this task, the $i^\text{th}$ datum is a graph with $N_i$ nodes represented by a pair $\{ \A_i, \X_i \}$ and must be associated to the correct label $\mathbf{y}_i$.
We test the models on different graph classification datasets.
For featureless graphs, we used the node degree information and the clustering coefficient as surrogate node features. 
We evaluate model performance with a 10-fold train/test split, using $10\%$ of the training set in each fold as validation for early stopping.
We adopt a fixed network architecture and only switch the pooling layers to compare Graclus, NDP, Top-$K$, SAGPool~\citep{lee2019selfattention}, DiffPool, and \mincutpool{}. 
All pooling methods are configured to drop half of the nodes in a graph at each layer.
As additional baselines, we consider the popular Weisfeiler-Lehman (WL) graph kernel~\citep{shervashidze2011weisfeiler}; a network with only MP layers (\textit{No-pool}), to understand whether or not pooling is useful for a particular task; a fully connected network (\textit{Dense}), to quantify how much additional information is brought by the graph structure w.r.t.\ the node features alone.

Tab.~\ref{tab:graph_class} reports the classification results. \mincutpool{} consistently achieves equal or better results with respect to every other method.
On the other hand, some pooling procedures do not always improve the performance compared to the \textit{No-pool} baseline, making them not advisable to use in some cases.
Interestingly, in some datasets such as Proteins and COLLAB it is possible to obtain fairly good classification accuracy with the \textit{Dense} architecture, meaning that the graph structure only adds limited information. We also note the good performance of the WL kernel on Mutagenicity.

%%%%%%%%%%%%%%%%%%%%%%%%%%%%%%%%%%%%%%%%%%%%%%%%%%%
%% CONCLUSIONS
%%%%%%%%%%%%%%%%%%%%%%%%%%%%%%%%%%%%%%%%%%%%%%%%%%%
\section{Conclusions}
We introduced a deep learning approach to address important limitations of spectral clustering algorithms, and designed a pooling method for graph neural networks that overcomes several drawbacks of existing pooling operators.
Our approach combines the desirable properties of graph-theoretical approaches with the adaptive capability of learnable methods. 
We tested the effectiveness of our method on unsupervised node clustering tasks, as well as supervised graph classification tasks on several popular benchmark datasets.
Results show that \mincutpool{} performs significantly better than existing pooling strategies for GNNs.

%%%%%%%%%%%%%%%%%%%%%%%%%%%%%%%%%%%%%%%%%%%%%%%%%%%
%% ACK & REFERENCES
%%%%%%%%%%%%%%%%%%%%%%%%%%%%%%%%%%%%%%%%%%%%%%%%%%%
\subsection*{Acknowledgments}
This research is funded by the Swiss National Science Foundation project 200021 172671: “ALPSFORT: A Learning graPh-baSed framework FOr cybeR-physical sysTems.”

\bibliographystyle{icml2020}
\bibliography{references}

%%%%%%%%%%%%%%%%%%%%%%%%%%%%%%%%%%%%%%%%%%%%%%%%%%%
%% APPENDIX
%%%%%%%%%%%%%%%%%%%%%%%%%%%%%%%%%%%%%%%%%%%%%%%%%%%
\onecolumn

\appendix
\appendixpage

\section{Normalized cut objective: trace ratio vs.\ ratio trace}

The normalized cut objective in spectral clustering is formulated as a trace of the ratio of two matrices, while the cut loss term $\mathcal{L}_c$ in the loss function proposed in this paper is a ratio of two matrix traces.
While the numerical values of these quantities are similar, there are some differences between the two formulations that are worth discussing.

The cluster assignment matrix $\S$ should reflect some properties in the data described by a positive semidefinite matrix $\A$, while avoiding some of the properties described by a second positive semidefinite matrix $\mathbf{B}$. 
The optimal solution can be obtained by finding a matrix $\S$ that maximizes the term $tr(\S^T\A\S)$, while minimizing at the same time $tr(\S^T\mathbf{B}\S)$, which represents a penalty term.
In the case of the normalized cut, $\mathbf{B}$ is the matrix of vertex degrees $\D$, which acts as a scale normalization constraint.
A similar formulation can be found in other graph embedding algorithms, such as Linear Discriminant Analysis, Marginal Fisher Information, and Local Discriminant Embedding.
A natural way of expressing the objective in these algorithms is through the \textit{trace ratio}: 

\begin{equation}
    \label{eq:ratio_trace}
    \argmax_\S \frac{tr(\S^T\A\S)}{tr(\S^T\D\S)}.
\end{equation}

The optimization problem in \eqref{eq:ratio_trace} is non-convex and, therefore, does not admit a close-form solution.
For this reason, the trace ratio is often approximated by a \textit{ratio trace} problem: 

\begin{equation}
    \label{eq:trace_ratio}
    \argmax_\S tr\left(\frac{\S^T\A\S}{\S^T\D\S}\right) = tr\left[ (\S^T\D\S)^{-1}(\S^T\A\S) \right].
\end{equation}

The simplified ratio trace formulation yields an optimization problem that is convex and admits a closed-form solution, which can be efficiently found trough the generalized eigenvalue decomposition.
This is exactly the approach followed by the traditional spectral clustering algorithm.

However, in a neural network the parameters are optimized with gradient descent rather than by relying on the closed-form solution.
In particular, in the proposed architecture the matrix $\S$ is the output of an MLP rather than the eigenvectors of the Laplacian, as in traditional spectral clustering.
Therefore, since the convexity of the optimization problem is not exploited in our framework, we optimize the original trace ratio objective in \eqref{eq:ratio_trace} instead of the ratio trace in \eqref{eq:trace_ratio}.
% In fact, the latter is an approximation whose purpose is to allow a spectral decomposition, which is anyway not used in the proposed framework, but its solution deviates from the original trace ratio objective and it is invariant under any non-singular transformation, which introduces uncertainty in the clustering result.
Indeed, the solution of \eqref{eq:trace_ratio} deviates from the original objective and it is also invariant under any non-singular transformation, which, in return, introduces uncertainty in the clustering result.
Additionally, the ratio trace formulation implies an inversion of a $K \times K$ matrix, $(\S^T\D\S)^{-1}$, which must be computed at every forward pass to evaluate the loss function; this is, of course, much less efficient than computing the trace ratio in the loss term $\mathcal{L}_c$.

\section{Additional Experiments}
\subsection{Graph Regression of Molecular Properties on QM9}

The QM9 chemical database is a collection of $\approx$135k small organic molecules, associated to continuous labels describing several geometric, energetic, electronic, and thermodynamic properties.\footnote{\url{http://quantum-machine.org/datasets/}}
Each molecule in the dataset is represented as a graph $\{ \A_i, \X_i \}$, where atoms are associated to nodes, and edges represent chemical bonds. The atomic number of each atom (one-hot encoded; C, N, F, O) is taken as node feature and the type of bond (one-hot encoded; single, double, triple, aromatic) can be used as edge attribute. In this experiment, we ignore the edge attributes in order to use all pooling algorithms without modifications. 

The purpose of this experiment is to compare the trainable pooling methods also on a graph regression task, but it must be intended as a proof of concept.
In fact, the graphs in this dataset are extremely small (the average number of nodes is 8) and, therefore, a pooling operation is arguably not necessary.
We consider a GNN with architecture \textit{MP(32)-pool-MP(32)-GlobalAvgPool-Dense}, where \textit{pool} is implemented by Top-$K$, Diffpool, or \mincutpool{}. The network is trained to predict a given chemical property from the input molecular graphs. 
Performance is evaluated with a $10$-fold cross-validation, using $10\%$ of the training set for validation in each split.
The GNNs are trained for 50 epochs, using Adam with learning rate 5e-4, batch size 32, and ReLU activations. We use the mean squared error (MSE) as supervised loss. 

The MSE obtained on the prediction of each property for different pooling methods is reported in Tab.~\ref{tab:gr_res}.
As expected, the flat baseline with no pooling operation (\textit{MP(32)-MP(32)-GlobalAvgPool-Dense}) yields a lower error in most cases.
Contrarily to the graph classification and the AE task, Top-$K$ achieves better results than Diffpool in average.
Once again, \mincutpool{} significantly outperforms the other methods on each regression task and, in one case, also the flat baseline.

% :::::::::::::::::::::::::: TAB GRAPH REG ::::::::::::::::::::::::::
\begin{table}[h!]
\setlength\tabcolsep{.7em} %horizontal padding
\small
\centering
\bgroup
\def\arraystretch{1} %vertical padding
\begin{tabular}{lccc|c}
\cmidrule[1.5pt]{1-5}
\textbf{Property} & \textbf{Top-$K$} & \textbf{Diffpool} & \textbf{\mincutpool} & \textbf{Flat baseline} \\
\midrule
mu       & 0.600\tiny{$\pm0.085$} & 0.651\tiny{$\pm0.026$} & \underline{\textbf{0.538}}\tiny{$\pm0.012$} & 0.559\tiny{$\pm0.007$}  \\
alpha    & 0.197\tiny{$\pm0.087$} & 0.114\tiny{$\pm0.001$} & \underline{0.078}\tiny{$\pm0.007$} & \textbf{0.065}\tiny{$\pm0.006$} \\
homo     & 0.698\tiny{$\pm0.102$} & 0.712\tiny{$\pm0.015$} & \underline{0.526}\tiny{$\pm0.021$} & \textbf{0.435}\tiny{$\pm0.013$} \\
lumo     & 0.601\tiny{$\pm0.050$} & 0.646\tiny{$\pm0.013$} & \underline{0.540}\tiny{$\pm0.005$} & \textbf{0.515}\tiny{$\pm0.007$} \\
gap      & 0.630\tiny{$\pm0.044$} & 0.698\tiny{$\pm0.004$} & \underline{0.584}\tiny{$\pm0.007$} & \textbf{0.552}\tiny{$\pm0.008$} \\
r2       & 0.452\tiny{$\pm0.087$} & 0.440\tiny{$\pm0.024$} & \underline{0.261}\tiny{$\pm0.006$} & \textbf{0.204}\tiny{$\pm0.006$} \\
zpve     & 0.402\tiny{$\pm0.032$} & 0.410\tiny{$\pm0.004$} & \underline{0.328}\tiny{$\pm0.005$} & \textbf{0.284}\tiny{$\pm0.005$} \\
u0\_atom & 0.308\tiny{$\pm0.055$} & 0.245\tiny{$\pm0.006$} & \underline{0.193}\tiny{$\pm0.002$} & \textbf{0.163}\tiny{$\pm0.001$} \\
cv       & 0.291\tiny{$\pm0.118$} & 0.337\tiny{$\pm0.018$} & \underline{0.148}\tiny{$\pm0.004$} & \textbf{0.127}\tiny{$\pm0.002$} \\
\cmidrule[1.5pt]{1-5}
\end{tabular}
\egroup
\caption{MSE on the graph regression task. The best results with a statistical significance of $p < 0.05$ are highlighted: the best overall are in bold, the best among pooling methods are underlined.}
\label{tab:gr_res}
\end{table}

\section{Experimental Details}

All GNN architectures in this work have been implemented with the Spektral library.\footnote{\url{https://graphneural.network}}
The code to reproduce all experiments is available at \url{https://github.com/FilippoMB/Spectral-Clustering-with-Graph-Neural-Networks-for-Graph-Pooling}.
For the WL kernel, we used the implementation provided in the GraKeL library.\footnote{\url{https://ysig.github.io/GraKeL/dev/}}
The pooling strategy based on Graclus, is taken from the ChebyNets repository.\footnote{\url{https://github.com/mdeff/cnn_graph}}

\subsection{Clustering on Citation Networks}
Diffpool and \mincutpool are configured with 16 hidden neurons with linear activations in the MLP and MP layer, respectively used to compute the cluster assignment matrix $\S$.
The MP layer used to compute the propagated node features $\X^{(1)}$ uses an ELU activation in both architectures.
The learning rate for Adam is 5e-4, and the models are trained for 10000 iterations. 
The details of the citation networks dataset are reported in Tab.~\ref{tab:cit_dataset}.

% :::::::::::::::::::::::::::: TAB. DATASET CLUST ::::::::::::::::::::::::::::
\bgroup
\def\arraystretch{1} %vertical padding
\setlength\tabcolsep{.5em} %horizontal padding
\begin{table}[!ht]
\footnotesize
\centering
\caption{Details of the citation networks datasets} 
\label{tab:cit_dataset}
\begin{tabular}{lcccc}
\cmidrule[1.5pt]{1-5}
\textbf{Dataset} & \textbf{Nodes} & \textbf{Edges} & \textbf{Node features} & \textbf{Node classes} \\
\cmidrule[.5pt]{1-5}
Cora   & 2708 & 5429 & 1433 & 7 \\
Citeseer & 3327 & 9228 & 3703 & 6 \\
Pubmed & 19717 & 88651 & 500 & 3 \\
\cmidrule[1.5pt]{1-5}
\end{tabular}
\end{table}
\egroup

\subsection{Graph Classification}
We train the GNN architectures with Adam, an L\textsubscript{2} penalty loss with weight 1e-4, and 16 hidden units ($H$) both in the MLP of \mincutpool{} and in the internal MP of Diffpool.
\textit{Mutagenicity}, \textit{Proteins}, \textit{DD}, \textit{COLLAB}, and \textit{Reddit-2k} are datasets representing real-world graphs and are taken from the repository of benchmark datasets for graph kernels.\footnote{\url{ https://ls11-www.cs.tu-dortmund.de/staff/morris/graphkerneldatasets}}
\textit{Bench-easy} and \textit{Bench-hard}\footnote{\url{ https://github.com/FilippoMB/Benchmark_dataset_for_graph_classification}} are datasets where the node features $\X$ and the adjacency matrix $\A$ are completely uninformative if considered alone. Hence, algorithms that account only for the node features or the graph structure will fail to classify the graphs. Since \textit{Bench-easy} and \textit{Bench-hard} come with a train/validation/test split, the 10-fold split is not necessary to evaluate the performance.
The statistics of all the datasets are reported in Tab.~\ref{tab:graph_dataset}.

% :::::::::::::::::::::::::::: TAB. DATASET CLASS ::::::::::::::::::::::::::::
\bgroup
\def\arraystretch{1} %vertical padding
\setlength\tabcolsep{.5em} %horizontal padding
\begin{table}[!ht]
\footnotesize
\centering
\caption{Summary of statistics of the graph classification datasets} 
\label{tab:graph_dataset}
\begin{tabular}{lcccccc}
\cmidrule[1.5pt]{1-7}
\textbf{Dataset} & \textbf{samples} & \textbf{classes} & \textbf{avg. nodes} & \textbf{avg. edges} & \textbf{node attr.} & \textbf{node labels} \\
\cmidrule[.5pt]{1-7}
Bench-easy   & 1800  & 3  & 147.82 & 922.66 & -- & yes \\
Bench-hard   & 1800  & 3  & 148.32 & 572.32 & -- & yes \\
Mutagenicity & 4337  & 2  & 30.32  & 30.77  & -- & yes \\
Proteins     & 1113  & 2  & 39.06  & 72.82  & 1  & no  \\
DD           & 1178  & 2  & 284.32 & 715.66 & -- & yes \\
COLLAB       & 5000  & 3  & 74.49  & 2457.78& -- & no  \\
Reddit-2K    & 2000  & 2  & 429.63 & 497.75 & -- & no \\
\cmidrule[1.5pt]{1-7}
\end{tabular}
\end{table}
\egroup

\section{Architectures Schemata}
\begin{figure}[!ht]
\centering
	\includegraphics[width=0.25\columnwidth]{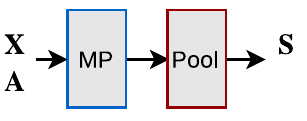}
    \caption{Architecture for clustering/segmentation.}
	\label{fig:scheme_clust}
	\vspace{1cm}

	\includegraphics[width=0.75\columnwidth]{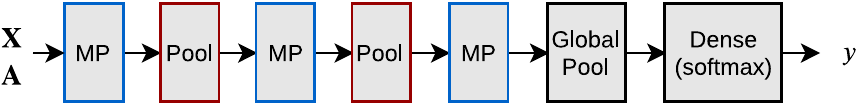}
    \caption{Architecture for graph classification.}
	\label{fig:scheme_class}
	\vspace{1cm}
	
	\includegraphics[width=0.55\columnwidth]{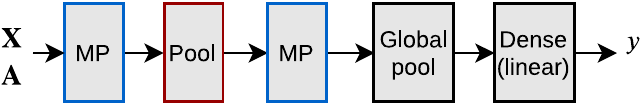}
    \caption{Architecture for graph regression.}
	\label{fig:scheme_reg}
	\vspace{1cm}
	
	\includegraphics[width=0.75\columnwidth]{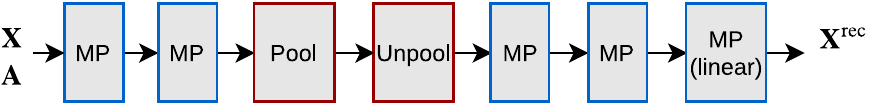}
    \caption{Architecture for the autoencoder.}
	\label{fig:scheme_ae}
\end{figure}

Fig.~\ref{fig:scheme_clust} depicts the GNN architecture used in the clustering and segmentation tasks; Fig.~\ref{fig:scheme_class} depicts the GNN architecture used in the graph classification task; Fig.~\ref{fig:scheme_reg} depicts the GNN architecture used in the graph regression task; Fig.~\ref{fig:scheme_ae} depicts the graph autoencoder used in the graph signal reconstruction task.

% \begin{figure}[!ht]
% \centering
% 	\includegraphics[width=0.55\columnwidth]{figs/regression_scheme.pdf}
%     \caption{Architecture for graph regression.}
% 	\label{fig:scheme_reg}
% 	\vspace{1cm}
	
% 	\includegraphics[width=0.75\columnwidth]{figs/autoencoder_scheme.pdf}
%     \caption{Architecture for the autoencoder.}
% 	\label{fig:scheme_ae}
% \end{figure}

% \end{document}

\end{document}